\documentclass[11pt,a4paper]{article}
\usepackage{times,latexsym}
\usepackage{multirow}
\usepackage{amssymb}
\usepackage{amsmath}
\usepackage{pifont}
\usepackage{microtype}
\usepackage{hyperref}
\usepackage{url}
\usepackage{booktabs}
\usepackage{amsfonts}
\usepackage{xcolor}         % colors
\usepackage{graphicx}
\usepackage{url}
\usepackage[T1]{fontenc}

\usepackage[acceptedWithA]{tacl2021v1}
\usepackage{xspace,mfirstuc,tabulary}
\usepackage{listings}

\newcommand{\inquotes}[1]{{``#1''}}
\newcommand{\dataset}{\emph{EvocativeLines}}
\newcommand{\mustd}[2]{$\text{#1}^{\text{(#2)}}$}
\newcommand{\mustdbf}[2]{$\mathbf{\textbf{{#1}}^{\textbf{(#2)}}}$}
\newcommand{\aid}[1]{$\mathcal{A}_{#1}$}
\newcommand{\highlight}[1]{\colorbox{gray!30}{#1}}
\newcommand{\s}[1]{$\mathbf{S_{#1}}$}

\definecolor{codegreen}{rgb}{0,0.6,0}
\definecolor{codegray}{rgb}{0.5,0.5,0.5}
\definecolor{codepink}{RGB}{252, 142, 172}
\definecolor{codepurple}{rgb}{0.58,0,0.82}
\definecolor{backcolour}{RGB}{245,245,245}
\lstdefinestyle{mystyle}{
    backgroundcolor=\color{backcolour},   
    commentstyle=\color{magenta},
    keywordstyle=\color{blue},
    numberstyle=\tiny\color{codegray},
    stringstyle=\color{codepurple},
    basicstyle=\fontfamily{\ttdefault}\footnotesize,
    breakatwhitespace=false,         
    breaklines=true,                 
    keepspaces=true,    
    frame=single,
    numbersep=5pt,                  
    showspaces=false,                
    showstringspaces=false,
    showtabs=false,                  
    tabsize=2,
    classoffset=1, % starting new class
    keywordstyle=\color{violet},
    classoffset=0,
}
\lstdefinelanguage{JavaScript}{
  keywords={typeof, new, true, false, catch, function, return, null, catch, switch, var, if, in, while, do, else, case, break},
  keywordstyle=\color{blue}\bfseries,
  ndkeywords={class, export, boolean, throw, implements, import, this},
  ndkeywordstyle=\color{darkgray}\bfseries,
  identifierstyle=\color{black},
  sensitive=false,
  comment=[l]{//},
  morecomment=[s]{/*}{*/},
  commentstyle=\color{purple}\ttfamily,
  stringstyle=\color{red}\ttfamily,
  morestring=[b]',
  morestring=[b]''
}

\newif\iftaclinstructions
\taclinstructionsfalse % AUTHORS: do NOT set this to true
\iftaclinstructions
\renewcommand{\confidential}{}
\renewcommand{\anonsubtext}{(No author info supplied here, for consistency with
TACL-submission anonymization requirements)}
\newcommand{\instr}
\fi

\iftaclpubformat % this ''if'' is set by the choice of options

\else

\fi

\title{Computational Modeling of Artistic Inspiration: A Framework for Predicting Aesthetic Preferences in Lyrical Lines Using Linguistic and Stylistic Features}

\author{
  Gaurav Sahu$^\diamond$ 
  \and
  Olga Vechtomova$^{\diamond \dagger}$
  \\
  \ \\
  $^\diamond$David R. Cheriton School of Computer Science \\
  $^\dagger$Department of Management Science and Engineering \\
  University of Waterloo, Canada \\
  \texttt{\{gaurav.sahu, olga.vechtomova\}@uwaterloo.ca}
  \\
}

\date{}

\begin{document}
\maketitle
\begin{abstract}
  % Artistic inspiration is one of the least understood aspects of human creativity, yet it is crucial for the creation of works that resonate deeply with audiences. The complexity and unpredictability of aesthetic stimuli and the vividness of poetic imagery are often key to evoking this inspiration.
  Artistic inspiration remains one of the least understood aspects of the creative process. It plays a crucial role in producing works that resonate deeply with audiences, but the complexity and unpredictability of aesthetic stimuli that evoke inspiration have eluded systematic study.
  This work proposes a novel framework for computationally modeling artistic preferences in different individuals through key linguistic and stylistic properties, with a focus on lyrical content.
  In addition to the framework, we introduce \textit{EvocativeLines}, a dataset of annotated lyric lines, categorized as either ``inspiring'' or ``not inspiring,'' to facilitate the evaluation of our framework across diverse preference profiles.
  Our computational model leverages the proposed linguistic and poetic features and applies a calibration network on top of it to accurately forecast artistic preferences among different creative individuals.
  % Our computational model leverages proposed linguistic and poetic properties to predict the artistic preference of an individual for a given line.
  Our experiments demonstrate that our framework outperforms an out-of-the-box LLaMA-3-70b, a state-of-the-art open-source language model, by nearly 18 points.
  Overall, this work contributes an interpretable and flexible framework that can be adapted to analyze any type of artistic preferences that are inherently subjective across a wide spectrum of skill levels.
\end{abstract}

\section{Introduction}
The creative process, along with the inspiration that drives the creation of art—whether visual, poetic, or musical—remains one of the least understood aspects of human experience. Every artist employs different methods to summon inspiration or, as the XIX century American artist Robert Henri noted, to ``be in that wonderful state which makes art inevitable.''

\citet{rubin2023creative} describes three distinct stages to a creative process: the Seed phase, where one is entirely open to any and all forms of inspiration; the Experimentation phase, where one tries multiple approaches to allow the nascent ideas or ``seeds'' to grow and take shape; and the Crafting phase, where the artist refines these ideas into a cohesive work.
This work focuses on the Seed phase, exploring how AI-generated poetic lines can inspire new creative pathways for artists.
Much like organically encountered stimuli, these AI-generated lines serve not as direct inputs into the final artwork but as triggers that guide the artist into a creative mindset.

Assessing creativity in AI-generated content is challenging due to its subjective and individual nature. Traditional evaluation methods, such as the Consensual Assessment Technique (CAT) \citep{amabile1982social} and the Torrance Test of Creative Thinking (TTCT) \citep{torrance1966torrance}, provide structured evaluation frameworks, but they may not be well-suited for evaluating AI-generated outputs intended to inspire artists rather than serve as final artifacts.
For instance, CAT uses collective judgement of experts to assess creativity, but it does not assess the inspirational potential or the ability of the outputs to spark further creative development in individual artists. Similarly, while TTCT measures divergent thinking, it has faced criticism for not reflecting the diverse forms that creativity can take \citep{baer2011divergent}.
% Overall, in the context of creating art, it is crucial to measure how well AI systems support individual creative processes that encompass a wide spectrum of artistic preferences and skill levels.

In this work, we propose a novel evaluation framework that identifies and formalizes key linguistic, and poetic properties of AI-generated lines that best explain the subjective preferences of artists during the Seed phase of the creative process.
Our framework does not assume a pre-requisite skill level of an artist and can be applied to study the artistic preferences of a wide spectrum of creative individuals for any task that is inherently subjective.
Further, we focus on analyzing the inspirational quality of lyric lines in the Seed phase rather than evaluating them as the final artifact, thereby addressing the key limitation of previous creativity evaluation frameworks that were too constrained in their definition of what they considered creative or inspiring.
% focusing on qualities that facilitate inspiration, enhance creative flow, and promote emotional engagement in creative individuals.

We test our framework on a collection of 3025 poetic lines generated by LyricJam~\cite{vechtomova2021lyricjam}, a publicly available research system that creates poetic lines
% in response to instrumental music,
with the goal of assisting creative individuals in finding novel lyrical and poetic ideas for use in the Seed phase of the creative process.
We refer to the dataset of poetic lines as {\dataset} and conduct comprehensive experiments to establish the reliability of our framework in accurately forecasting the artistic preferences of an individual.
Lastly, we provide insights about the suitability of recent large language models (LLMs) like LLaMA-3-70b~\cite{dubey2024llama} for our task of modeling diverse artistic preferences, which involves a high degree of subjectivity.

To summarize our contributions:
    \textbf{1)} we propose a novel framework for evaluating AI-generated poetic lines that focuses on identifying and formalizing key linguistic and poetic properties that explain artists' subjective preferences during the early creative process (Seed phase);
    \textbf{2)} our framework is highly interpretable and is designed to be applicable across a wide range of creative individuals, regardless of their skill level;
    % It specifically addresses the inspirational quality of lyric lines in the early stages of creation.
    \textbf{3)} we conduct comprehensive experiments to establish the framework's reliability in predicting artistic preferences over multiple strong baselines like LLaMA-3-70b on a large dataset of AI-generated poetic lines (referred to as {\dataset}); and
    \textbf{4)} we provide insights into the suitability of recent large language models, such as LLaMA-3-70b, for modeling diverse artistic preferences in highly subjective tasks.
% We obtain multiple human annotations for the lines and use those annotations to determine the suitability of our framework inspiring lines and determine the extent to which recent large language models are suitable for computational modeling of artistic preferences.

\section{Related Work}
\label{sec:related}
% \paragraph{LLM-based Evaluation of Natural Language Texts:}
Recent advancements in natural language processing (NLP), particularly in the space of LLMs, have facilitated the development of evaluation metrics, such as LLM-Rubric~\citep{hashemi2024llm}, GPTScore~\citep{fu-etal-2024-gptscore}, and G-Eval~\citep{liu-etal-2023-g}, that are far better aligned with human judgements than traditional NLP metrics like BLEU~\citep{papineni2002bleu}, ROUGE~\citep{lin2004rouge}, or METEOR~\citep{banerjee2005meteor}; however, these LLM-based metrics primarily assess general language quality rather than creative attributes.
There are also evaluation approaches that measure aspects like diversity and fluency but do not capture the multifaceted nature of creative expression.
For instance, \citet{Hashimoto2019Unifying} proposed an evaluation framework balancing diversity and fluency, while \citet{pillutla2021mauve} introduced MAUVE, an automatic measure of how close generated text is to human text.

% , which is highly subjective and can vary from one person to another.
% In comparison, our work focuses on developing a framework that can be adapted to reflect the subjective preferences of different creative individuals.
% \paragraph{AI Systems for Subjective NLP Tasks:}
% ~\citet{hessel2022androids} propose a humour detection benchmark that evaluates the ability of AI models to ``understand" humour, but they heavily rely on human annotations for evaluation.

Recently, \citet{chakrabarty2024art} extended the TTCT framework to the Torrance Test of Creative Writing (TTCW) to evaluate creativity in outputs from both humans and large language models (LLMs), but it also suffers from TTCT's limitation of being unable to capture the diverse forms creativity can take.
In our work, we develop multidimensional evaluation metrics that capture various linguistic and stylistic characteristics of generated text and we also investigate how these metrics can predict personal preferences of individuals with diverse artistic skills and interests. 
% engaged in creative pursuits that transcend traditional genres and fields.
    
\section{The Proposed Framework}
% \paragraph{Motivation.}
The relationship between aesthetic pleasure and complexity, known as the inverted U-shape or ``Wundt'' effect, suggests that stimuli of intermediate complexity are most pleasing. This U-shape effect, first explored by Berlyne~\shortcite{Berlyne1975}, has been observed in various domains such as music \cite{Gold9397} and product design \cite{Althuizen2021}.
%\citet{Gold9397} offered a possible explanation for the inverted U-shaped effect, grounded in learning. Learning engages dopaminergic reward system in the brain, and when the events (in our case, the sequence of words in a sentence) are completely predictable, little or no learning is taking place.
%On the other end of the complexity scale, when the events are fully random, no patterns can be discerned, hence no learning can take place either. However, events with intermediate degrees of predictability enhance learning and make us engage with the stimulus more meaningfully.
% Complexity in works of art, such as music and poetry, is often also associated with unpredictability and surprise.
Many artists and poets since the early 20th century have recognized that established language forms can become worn out and lose their evocative power. William Burroughs argued that language acts as a "lock" that restricts creativity and confines us to predictable patterns~\cite{burroughs2012}.
Burroughs popularized the cut-up technique, which involves creating poetic texts by randomly re-arranging words. This method was later adopted by music artists like David Bowie~\cite{jones2017david} and Kurt Cobain~\cite{cross2001heavier}, who used it to find inspiration and generate new creative ideas for their lyrics. Poets and writers have also sought other ways to make their writing less predictable. Tom Spanbauer, for instance, coined the term "burnt tongue" to describe a literary device where language is deliberately misused to slow readers down and force them to engage more deeply with the text~\cite{palahniuk2020consider}. This intuition is supported by eyetracking studies, which have shown that predictable words are often skipped or elicit shorter fixation times~\cite{Lowder2018}.
While there is evidence that the U-shape effect exists across different aesthetic domains, there is also strong evidence that our preferences for aesthetic stimuli are highly individual~\cite{Gold9397, Althuizen2021}.
Inspired by the different theories and evidence, we propose a two-step framework that \textbf{a)} measures multiple poetic and linguistic, characteristics of a lyric line; and \textbf{b)}
uses those measurements to learn the artistic preferences of an individual.

\section{Step 1: Identifying The Key Poetic and Stylistic Characteristics}
\label{sec:features}
% In this section, we describe the linguistic devices and poetic structures that contribute to the evocative quality of AI-generated lyric lines, examining how these elements are perceived as inspiring and how their interaction stimulates artistic creativity during the initial Seed phase of the creative process. We also detail the construction of features that represent these linguistic and poetic aspects, explaining how they are quantified and incorporated into our model. \comment{Please check to see if this reflects your intent - OV}
\paragraph{Poetic/Stylistic Characteristics.}
There are over 100 stylistic devices in the English language that a poet or lyricist can use to achieve a certain emotional effect.
The most prominent stylistic devices in poetry includes:
\textbf{sound devices} such as alliteration and assonance
% , rhyme, and meter
focusing on the auditory properties of the poem; \textbf{figurative and descriptive language} like metaphor, and poetic imagery
% , symbolism, and diction
that encompasses the usage of language in non-literal ways to convey a deeper meaning or create vivid imagery; and \textbf{structural devices} like anaphora, and epistrophe
% , and parallelism
that relate to the arrangement of words and lines.
%alternative to consider: These devices often focus on different aspects of the poem:
%\begin{itemize} \item \textbf{Sound Devices} emphasize the auditory properties of the poem, such as alliteration, assonance, rhyme, and meter, contributing to the overall musicality and rhythm. \item \textbf{Figurative and Descriptive Language} involves the use of language in non-literal ways to convey deeper meaning or create vivid imagery. This includes devices like metaphor, simile, poetic imagery, symbolism, and diction. \item \textbf{Structural Devices} pertain to the arrangement of words and lines, such as anaphora, epistrophe, and parallelism, which determine the poem’s structure and impact. \end{itemize}
While many of these devices act together to create the desired effect, a poem is foremost intended to evoke emotion in the reader and heighten perceptual awareness~\cite{Dunnigan2014}. Figurative and descriptive language plays an important role in achieving this purpose. For this reason, we design the following four features to capture and reflect these critical aspects of poetry.

\subsection{Poetic Imagery}
\textit{Poetic Imagery} is a stylistic device that involves references to physical objects, appealing to the reader’s senses (e.g., visual, auditory, tactile) to evoke specific emotions or feelings or make abstract ideas more tangible and relatable~\citep{pound1913few,brooks2004warren,kao2012computational}.

To compute poetic imagery for a lyric line, we prompt a LLaMA-3-70b model to output a rating between 1-5.
Since poetic imagery is a well-established stylistic device in English, we perform prompting in a zero-shot fashion.
% and do not include any example human annotations.
This further avoids the induction of preference bias in the LLM outputs.
A higher rating denotes a higher degree of poetic imagery.
Formally, given the set of ratings $\{r_1, r_2,r_3,r_4,r_5\}$, we compute the LLM score for a sentence $S$ as follows:

\begin{equation}
    score_{img}(S) = \sum_{i=1}^{5} p(r_i) \times r_i,
    \label{eq:geval}
\end{equation}
where $p(r_i)$ is the probability of the LLM assigning an imagery rating of $r_i$ to the sentence $S$ (see Figure~\ref{fig:pi_prompt} in Appendix~\ref{app:prompts} for the full prompt).
We use a weighted summation instead of directly using the output tokens as they provide more fine-grained, continuous scores that better capture the nuances of texts~\citep{liu-etal-2023-g}.
To construct the final imagery feature vector, we combine the scores of all the subsequences in a given lyric line to also capture the temporal fluctuations and development in the sentence:
\begin{equation}
        \mathbf{S_{imagery}} = \langle score_{img}(\mathcal{S}_t) \rangle_{t=1}^{T},
        \label{eq:imagery}
\end{equation}
where $\mathcal{S}_t = (w_1,w_2,\cdots,w_t)$ denotes a subsequence with the first $t$ words in $S$.
We find this method to appropriately capture the poetic imagery of a lyric line.
For instance, ``there's a little red to the sea'' creates a more evocative image in the reader's mind and has a higher imagery score of 4.01 compared to ``this is all,'' which only achieves a score of 1.27.

\subsection{Word Energy}
\textit{Word Energy} is a composite feature that encompasses symbolism and diction, both of which contribute to the emotional and conceptual impact of a poem. Symbolism uses references to objects that evoke in the reader's mind certain abstract concepts, associations, and emotions~\citep{cassirer1946language}.
For example, the word ``chains'' is used to represent a feeling of entrapment and lack of control. Diction refers to the choice of words made by the poet to achieve a desired effect~\cite{brooks2004warren}. For example, by choosing the word ``beseeching'' over a more neutral word ``searching'', the poet emphasizes the sense of longing and desperation.
Similar to imagery, we prompt a LLaMA-3-70b model to output a rating between 1-5 to measure the notion of word energy in a given lyric line.
Note that the prompting is zero-shot, and since word energy is a composite feature, we include three example terms (not complete phrases) that reflect the notion of word energy.
A higher rating denotes higher word energy.
The final feature vector also assumes a similar form to Equation~\ref{eq:imagery} for a lyric line $S$:
\begin{equation}
        \mathbf{S_{energy}} = \langle score_{eng}(\mathcal{S}_t) \rangle_{t=1}^{T},
        \label{eq:energy}
\end{equation}
where $\mathcal{S}_t = (w_1,w_2,\cdots,w_t)$ denotes a subsequence with the first $t$ words in $S$, and $score_{eng}(\mathcal{S}_t)$ is computed similar to $score_{img}(\mathcal{S}_t)$ in Equation~\ref{eq:geval} except the ratings reflect the notion of word energy instead of poetic imagery.
For example, ``all the words will drown'' achieves a score of 4.11 compared to ``where i know about you,'' which gets a score of 1.92.

\subsection{Level of Abstraction} We measure the level of abstraction or ambiguity in a line, as a higher level of abstraction indicates that the line is more likely to be unresolved and allows for multiple interpretations; hence, it can be potentially more engaging.
The \textit{Linguistic Category Model (LCM)} is a widely employed theoretical framework that considers the social–cognitive functions of four linguistic categories and can measure the degree of abstraction in a sentence~\citep{semin1988cognitive,johnson2020measuring}.
As a feature for quantifying the degree of abstraction of the language used in a poetic line, this measure captures key stylistic devices, such as Diction, by assessing the specific types of words chosen: Descriptive Action Verbs (DAVs), Interpretative Action Verbs (IAVs), State Verbs (SVs), and Adjectives (ADJs). LCM is intended to represent the balance between abstract and concrete language, which is important in defining the poem’s tone, mood, and the vividness of its imagery. Furthermore, the LCM indirectly indicates the presence of figurative language, such as metaphors and similes, by highlighting the use of more abstract expressions.
We compute the feature vector representing the level of abstraction of a sentence $S$ as follows:

    \begin{equation}
        \mathbf{S_{abs}} = \langle score_{abs}(\mathcal{S}_t) \rangle_{t=1}^{T}
        \label{eq:hLCM}
    \end{equation}
where $\mathcal{S}_t = (w_1,w_2,\cdots,w_t)$ denotes a subsequence with the first $t$ words in $S$, and $score_{abs}(\mathcal{S}_t)$ is computed as follows:

    \begin{equation}
        score_{abs} = \frac{D + (2 \times I) + (3 \times S) + (4 \times A)}{D + I + S + A}
        \label{eq:lcm}
    \end{equation}
    In Equation~\ref{eq:lcm}, $D, I, S$ and $A$ denote the number of times DAVs like ``eating'' and ``walking,'' IAVs like ``helping'' and ``playing,'' SVs like ``love'' and ``admire,'' and ADJs like ``ethical'' and ``uneven'' occur in the text.
    Weights assigned to the four counts are based on their theorized abstraction level determined through a linguistic study of 40,000 English words~\citep{brysbaert2014concreteness}.
    Here, definitive, concrete verbs (DAVs) receive the lowest weight of 1.0, whereas abstract adjectives get the highest weight of 4.0.

    We prompt a LLaMA-3-70b model to obtain the counts of the four linguistic categories in a lyric line.
    % Figure~\ref{fig:hLCM} (Appendix~\ref{app:prompts}) shows the complete prompt.
    We compute the final feature vector 
    
    We find our scores highly reliable for this task as they are able to distinguish between the phrase ``rearranging this stage,'' which is very definitive and has a lower level of abstraction compared to ``the greatest began to deny,'' which remains somewhat unresolved.
    Our experiments (discussed in Section~\ref{sec:results}) also indicate that the level of abstraction in a line is one of the most prominent features in identifying potentially evocative lines for an artist.
    
\subsection{Valence} \textit{Valence} reflects the emotions expressed in a lyric line and is crucial for assessing an artist's preferred emotional tone that may guide their own creative process~\citep{scherer1984nature,charland2005heat}.
% This measure directly relates to key stylistic devices such as tone and mood, capturing the emotional tone conveyed by the poetic line. By identifying the specific emotions expressed, Valence also indirectly measures the effectiveness of figurative language and poetic imagery in evoking the desired emotional response from the reader.
For instance, an artist might favor lines that creatively blend positive and negative connotations, such as ``the greatest began to deny,'' or they may prefer lines that convey a purely positive sentiment, like ``tranquility in the lovers.''

To measure the valence of a line, we build a computation model of emotion that assigns a valence score based on the detected emotional tone of a line.
\textbf{First,} we define a list of 29 fine-grained emotion categories--including admiration, nostalgia, gratitude, grief, and remorse.
The list of emotions is derived from the GoEmotions dataset~\citep{demszky2020goemotions}, with the addition of one extra category, ``nostalgia,'' to capture the unique emotional nuances often present in lyrical content.
\textbf{Next,} we employ a prompting-based classifier built on the LLaMA-3-70b language model to obtain the probabilities for the top 5 emotional categories for a given lyric line.
% (see Figure~\ref{fig:valence_clf} in Appendix~\ref{app:prompts}) for the complete emotion classification prompt).
Let $S$ represent the input sentence, and $E_k$ denote the set of the top-$k$ emotions predicted by the classifier (with $k=5$ in our case). The probability vector for $S$ is then constructed as follows:

\begin{equation}
\mathbf{p}(S) = \mathcal{T}(f(S)) = \langle \tilde{p}_1, \tilde{p}_2, \dots, \tilde{p}_m \rangle,
\label{eq:p_S}
\end{equation}

where $m$ is the total number of emotion categories ($m=29$ in our case), $f(\cdot)$ denotes the probabilities of the top 5 emotion categories predicted by the LLaMA-3.1 classifier, and $\mathcal{T}$ represents the transformation defined as:

\begin{equation}
    \tilde{p}_{e} = 
        \begin{cases}
        p_{e}, & \text{if } e \in E_k \\
        p_{rem}, & \text{if } e \notin E_k
        \end{cases}
\end{equation}
where $p_e$ is the probability of emotion $e$ predicted by the LLaMA classifier and $p_{rem} = \frac{1 - \sum_{e \in E_k} p_{e}}{m-k}$ represents the remaining probability mass, equally distributed among the categories not included in the top 5. In essence, this process preserves the probabilities of the top 5 emotions as predicted by LLaMA while redistributing the remaining probability mass evenly across the other emotion categories.
\textbf{Finally,} we construct the complete valence vector for a lyric line $S=(w_1, w_2,\dots, w_T)$ as follows:

\begin{equation}
        \mathbf{S_{vfull}} = \langle \mathbf{p}(\mathcal{S}_t) \rangle_{t=1}^{T}, 
        \label{eq:valence_full}
\end{equation}
where $\mathcal{S}_t = (w_1,w_2,\cdots,w_t)$ denotes a subsequence with the first $t$ words in $S$.
We use subsequences to capture the temporal development of valence.

We also construct a uni-dimensional variant of $\mathbf{S_{vfull}}$, where we first group the 29 fine-grained emotions into three sets $E_{+}, E_{0}, E_{-}$ denoting sets of positive, neutral, and negative emotions, respectively.
% collapse the probability vector in.
Then, we collapse the full emotion probability vector in Equation~\ref{eq:p_S} to an overall score $\tilde{p} \in [-1, 1]$ denoting the probability of a sentence conveying a positive emotion:
\begin{equation}
\mathbf{p_{bin}}(S) = \tilde{p}_{+} = \sum \mathbb{I}_{e} \cdot \tilde{p}_{e},
\end{equation}
where $\mathbb{I}_{e}$ is an indicator function defined as:
\begin{equation}
\mathbb{I}_{e} = \begin{cases}
    +1, & \text{if } e \in E_{+} \\
    0, & \text{if } e \in E_{0} \\
    -1, & \text{if } e \in E_{-}
\end{cases}
\end{equation}
so that the final feature vector becomes:
\begin{equation}
        \mathbf{S_{vbin}} = \langle \mathbf{p_{bin}}(\mathcal{S}_t) \rangle_{t=1}^{T},
        \label{eq:valence_bin}
\end{equation}

Notably, we also tried a RoBERTa classifier~\citep{liu2019roberta} fine-tuned on the GoEmotions dataset as the classifier $f(\cdot)$, but our preliminary results suggested that features from the LLaMA-based prompting-based classifier have higher predictive power.

\paragraph{Linguistic and Statistical Characteristics.}
In addition to quantifying the stylistic characteristics and emotional aspects, we measure the inherent complexity of a poetic line.
Studies in experimental aesthetics suggest that stimuli of intermediate complexity are most strongly associated with pleasure. Complexity is typically expressed in features such as predictability and surprise, and we propose the following features to measure them:

\subsection{Surprisal} \textit{Surprisal}~\citep{shannon1948mathematical} of a word indicates the predictability of that word in a sentence.
    It is measured as the negative log probability of the word given its context:
    \begin{equation}
        s (w_t) = - \log_2 P(w_t | \mathbf{w}_{<t}),
        \label{eq:surprisal_word}
    \end{equation}
    where, $s (w_t)$ denotes the surprisal value for the $t$-th word of the sentence.
    % Equation~\ref{eq:surprisal_word} computes surprisal for a single word.
    To compute the surprisal of the complete sentence, we compose a vector of word-level surprisals, appended with the average surprisal value of all the words in the sentence as follows:
    % \begin{equation}
    %     \mathbf{S_{surprisal}} = \begin{bmatrix}
    %         s(w_1) \\
    %         s(w_2) \\
    %         \vdots \\
    %         s(w_T) \\
    %         s_{avg}
    %     \end{bmatrix},
    % \end{equation}
    \begin{equation}
        \mathbf{S_{surprisal}} = \langle s(w_t) \rangle_{t=1}^{T} \oplus \langle \overline{s} \rangle,
    \label{eq:surprisal_sentence}
    \end{equation}
    where $T$ denotes the number of words in the sentence, $\overline{s} = \sum s(w_t) / T$ denotes the mean of surprisal values for all the words in the sentence, and $\oplus$ denotes the concatenation operation.
   Since we do not have access to the true probability distribution $P$, we use GPT-2~\citep{radford2019language}\footnote{specifically, we use the 774M parameter \texttt{gpt2-large} model from Huggingface: \url{https://huggingface.co/openai-community/gpt2-large}}, a powerful autoregressive language model pre-trained on a large text corpus, to obtain a good approximation of the desired log probabilities $P(w_t | \mathbf{w}_{<t})$.
   The described surprisal feature can distinguish between a sentence like ``a knowledge of the world,'' which has a lower surprisal ($s_{avg}=10.12$), indicating higher predictability, and ``a just dream with the fire,'' which has a higher surprisal ($s_{avg}=11.21$), indicating a higher potential of novelty in the sentence (note that these values are on a log scale, so the difference is exponential.)
    
\subsection{Contextual Entropy} \textit{Contextual Entropy}~\citep{shannon1948mathematical} of a word is defined as the expected value of its \textit{Surprisal}:
    \begin{equation}
        h(w_t) = \sum_{w_i \in V} p(w_t | \mathbf{w}_{<t}) \log_2 P (w_t | \mathbf{w}_{<t}),
    \end{equation}
    where $h(w_t)$ denotes the contextual entropy value for the $t$-th word of the sentence, and $V$ denotes the vocabulary (all possible words that can appear next).
    Similar to surprisal, we use GPT-2 to compute the probabilities and construct the final sentence-level entropy feature vector by combining the word-level entropy values and their mean as follows:
    % \begin{equation}
    %     \mathbf{S_{entropy}} = \begin{bmatrix}
    %         h(w_1) \\
    %         h(w_2) \\
    %         \vdots \\
    %         h(w_T) \\
    %         h_{avg}
    %     \end{bmatrix},
    %     \label{eq:entropy}
    % \end{equation}
    \begin{equation}
        \mathbf{S_{entropy}} = \langle h(w_t) \rangle_{t=1}^{T} \oplus \langle \overline{h} \rangle,
    \label{eq:entropy}
    \end{equation}
    where $T$ denotes the number of words in the sentence, and $\overline{h} = \sum h(w_t) / T$ denotes the mean of entropy values for all the words in the sentence.
    Like surprisal, contextual entropy can also help us gauge the unpredictability/novelty of a sentence.
    For instance, ``i remember you'' has a lower entropy ($h_{avg}=6.21$) compared to ``pure crime of the lost'' ($h_{avg}=9.53$).
    Additionally, contextual entropy also exhibits a strong positive correlation with the reading times of a word~\citep{Lowder2018}, indicating its potential to capture the complexity of a sentence as well (higher entropy $\rightarrow$ higher complexity).

\subsection{Normalized Pointwise Mutual Information (NPMI)} \textit{NPMI} measures the association of two words by comparing the probability of their co-occurrence with the probabilities of their individual occurrences in a corpus~\citep{bouma2009normalized}.
Formally, the NPMI of a word $x$ occurring next to a word $y$ is defined as follows:

    \begin{equation}
        N(x, y) = \left(\ln \frac{p(x, y)}{p(x) p(y)}\right) / {- \ln p(x, y)}
        \label{eq:npmi}
    \end{equation}
    In Equation~\ref{eq:npmi}, $p(x) = count(x)/N$, where $count(x)$ denotes the number of occurrences of the word $x$ in the corpus and $N$ is the total number of word occurrences in the corpus. Similar definitions hold for $p(y)$ and $p(x, y)$.
    NPMI value is bounded between $[-1, 1]$, where a high NPMI value denotes high predictability and vice versa.
    As an example, the pair (\inquotes{see}, ``you'') has a high NPMI value in our corpus ($0.47)$ compared to (``never'', ``the''), which has a low NPMI value ($-0.11$) as ``see'' and ``you'' often occur together, whereas ``never'' is more frequently followed by verbs like ``hurt,'' ``end,'' and ``leave'' compared to a determiner like ``the.''

    In our work, we compute two variants of NPMI values: unidirectional and bidirectional.
    When computing unidirectional $N(x,y)$, we only consider the co-occurrence of $(x,y)$ if $y$ appears \textit{immediately after} $x$, but for bidirectional $N(x,y)$, we also consider the co-occurrence if $y$ appears \textit{immediately before} $x$.
    We denote the unidirectional and bidirectional NPMI values as $N_{1} (x, y)$ and $N_{2} (x, y)$, respectively.
    We construct the final uni- and bi-directional NPMI vectors in a similar fashion to surprisal and contextual entropy:
    % \begin{equation}
    %     \mathbf{S_{npmi,uni}} = \begin{bmatrix}
    %         N_{uni}(w_1, w_2) \\
    %         N_{uni}(w_2, w_3) \\
    %         \vdots \\
    %         N_{uni}(w_{T-1}, w_T) \\
    %         N_{uniavg}
    %     \end{bmatrix}
    % \end{equation}
    % \begin{equation}
    %     \mathbf{S_{npmi,bi}} = \begin{bmatrix}
    %         N_{bi}(w_1, w_2) \\
    %         N_{bi}(w_2, w_3) \\
    %         \vdots \\
    %         N_{bi}(w_{T-1}, w_T) \\
    %         N_{biavg}
    %     \end{bmatrix},
    % \end{equation}
    \begin{equation}
        \mathbf{S_{{npmi,uni}}} = \langle N_{uni}(w_t, w_{t+1}) \rangle_{t=1}^{T-1} \oplus \langle \overline{N}_{{uni}}\rangle
        \label{eq:npmi_uni}
    \end{equation}
    \begin{equation}
        \mathbf{S_{npmi,bi}} = \langle N_{bi}(w_t, w_{t+1}) \rangle_{t=1}^{T-1} \oplus \langle \overline{N}_{bi} \rangle
    \label{eq:npmi_bi}
    \end{equation}
    where $T$ denotes the number of words in the sentence, and $\overline{N}_{uni} = \sum N_{uni}(w_t, w_{t+1}) / T$ denotes the mean of unidirectional NPMI values for all the word pairs in the sentence and $\overline{N}_{bi} = \sum N_{bi}(w_t, w_{t-1}) / T$ denotes the mean of bidirectional NPMI values.

\subsection{Banality}
\textit{Banality} refers to the use of common or clichéd phrases, themes, or expressions that lack originality or depth.
It can also reflect the cultural and temporal context of the lyrics. For instance, an artist may use clichéd expressions to resonate with a broad audience or fit a certain era.
Therefore, banality is a crucial feature to measure when studying the creative preferences of different artists.
Similar to poetic imagery and word energy, we prompt a LLaMA-3-70b model to provide a rating from 1-5 that indicates how banal or cliché a poetic expression is.
Again, we prompt in a zero-shot fashion, and
a high rating indicates a banal expression, whereas a low rating indicates a unique or original expression.
We construct the final feature vector denoting banality as follows:
\begin{equation}
        \mathbf{S_{banality}} = \langle score_{bnl}(\mathcal{S}_t) \rangle_{t=1}^{T},
        \label{eq:banality}
\end{equation}
where $\mathcal{S}_t = (w_1,w_2,\cdots,w_t)$ denotes a subsequence with the first $t$ words in $S$, and $score_{bnl}(\mathcal{S}_t)$ is computed similar to $score_{img}(\mathcal{S}_t)$ except the ratings now reflect the level of banality instead of poetic imagery or word energy.
For instance, the expression ``take the test'' is a banal expression and achieves a score of 3.18 compared to ``breathe back the world,'' which achieves a score of 0.98.
Table~\ref{tab:features} in Appendix~\ref{app:ds} summarizes the different features proposed in this section along with their notations.

\section{Step 2: Training the Calibration Network}
Artistic preferences can vary from one individual to another -- a lyric line with the same stylistic and linguistic properties might be inspiring to one person but not another.
Therefore, we must calibrate the measurements from the first step to an individual's artistic preference before we can predict the artistic preference of an individual for a given lyric line.

We train the calibration network on a classification objective: given the linguistic and poetic features of a lyric line, predict if the line is inspiring to the end user.
Specifically, consider a dataset of artistic preferences of a creative individual, $D = \{(x_1, y_1), (x_2, y_2),\dots, (x_n, y_n)\}$, where $x_i \in \mathcal{X}$ denotes features of the $i$-th lyric line in the dataset and $y_i \in \mathcal{Y}$ denotes the corresponding label of whether or not the artist likes the lyric line.
Then, the goal of the calibration network is to learn a function $f_{\theta}: \mathcal{X} \rightarrow \mathcal{Y}$ that maps each feature vector $x_i$ to a probability distribution over classes $\hat{y} = f_{\theta}(x_i) = P(y | x_i; \theta)$.
For our binary classification setup, $\mathcal{Y} = \{\textrm{inspiring}, \textrm{not inspiring}\}$.
Notably, we do not directly map the embeddings of lyric lines to the predictions during calibration, as sentence embeddings do not necessarily capture the creative attributes and focus more on the general semantic and syntactic properties of the text.

\section{Dataset}
\label{sec:dataset}
% We now describe our process of constructing a dataset to measure the predictive power of the proposed features in identifying inspiring lyric lines.
We collect 3,025 lyric lines from the LyricJam platform, a system that autonomously generates novel lyric lines using bimodal neural networks~\cite{vechtomova2021lyricjam,vechtomova2023lyricjam}~\footnote{\url{https://lyricjam.ai}}.
% Specifically, we log the lines generated on the LyricJam platform for XX hours \todo{replace XX}, which resulted in a total of 3,025 lines.
We do not perform any pre-processing on the lyric lines, but notably, all of them are lowercase.
Next, the first and the last author labeled each line as either ``inspiring'' or ``not inspiring.''
By ``inspiring,'' we refer to whether these lines created an emotional impact and stimulated created ideas.
Both the first and the last authors actively engage in the process of creating art--including writing poems and short stories--which makes them suitable for the labeling task.
We refer to our final dataset of 3,025 annotated lyric lines as the `Evocative Lyric Lines Dataset' (or {\dataset} for short.)
The labeling process took approximately 12 hours, spread over multiple weeks.

To test the robustness of our framework on more preference profiles, we reached out to eight human labelers, each actively engaging in at least one creative activity like writing poetry or playing a musical instrument. %, either as a hobbyist or an amateur.
Specifically, we asked each annotator to rate a lyric line on a scale of 1-10 based on how inspiring they found the line.
Figure~\ref{fig:user_study_txt} shows the complete guidelines presented to the annotators, where we also clarify what we mean by `inspiration.'
We perform mix-max scaling to normalize the annotator ratings and obtain ``inspiring'' and ``not inspiring'' ratings.
Notably, we show the eight human annotators a subset of 200 lyric lines from the original dataset to keep the labeling process fast and efficient.
We obtain the 200 lines by randomly selecting 100 ``inspiring'' and 100 ``not inspiring'' lines of varying lengths (short, mid, and long) from the original dataset.
Choosing a smaller subset ensured that the estimated labeling time remained manageable, ranging from 30 to 45 minutes, thus also minimizing decision fatigue among the annotators.
This strategy allowed us to obtain highly reliable annotations while covering a broad spectrum of creative preferences to test the framework.

\begin{figure}
    \centering
    \includegraphics[width=\linewidth]{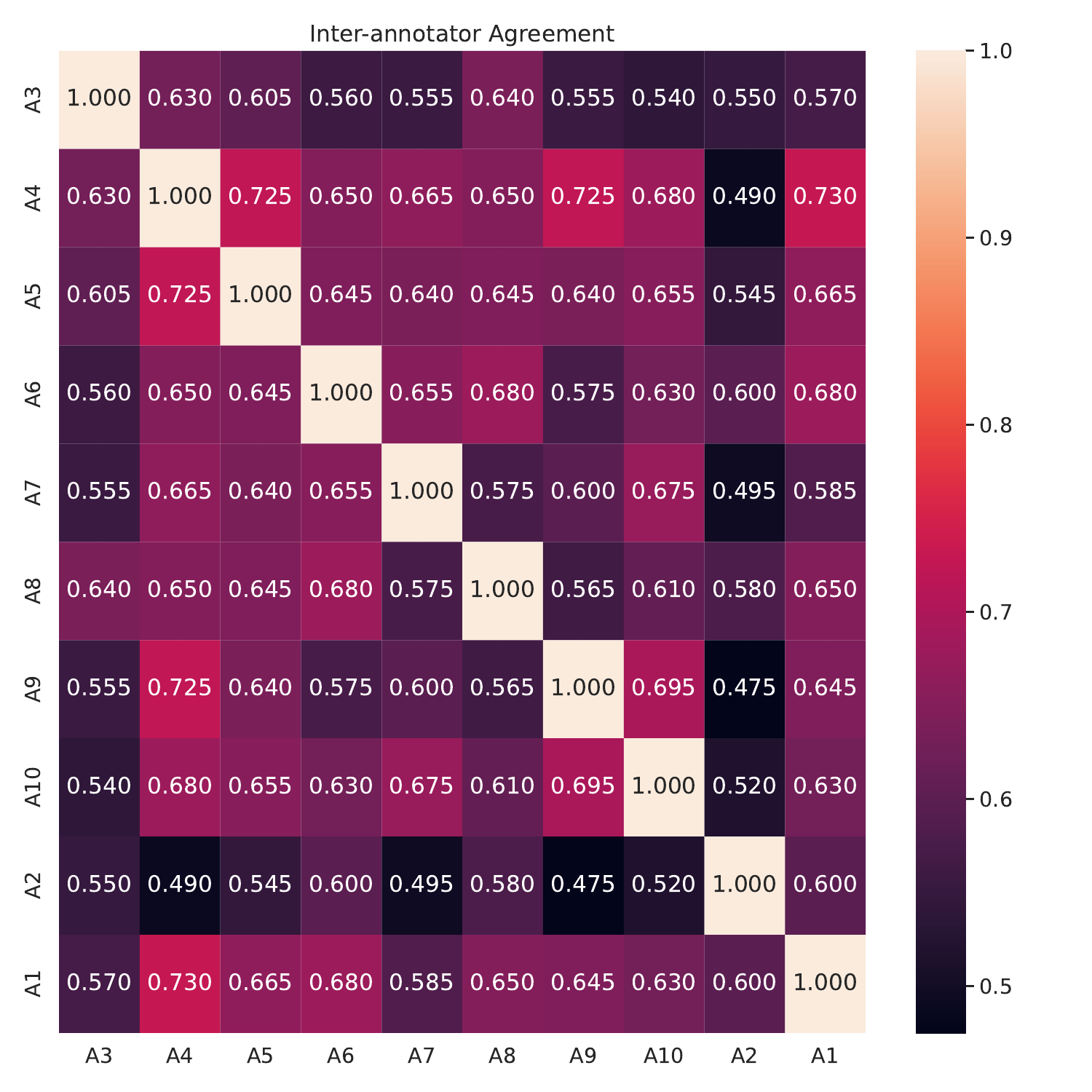}
    \caption{Inter-annotator agreement on the \textbf{EvocativeLines (small)} dataset.}
    \label{fig:cm}
\end{figure}

We denote our set of annotators as $\mathcal{A} = \{\mathcal{A}_1, \mathcal{A}_2,\cdots,\mathcal{A}_{10}\}$, where $\mathcal{A}_1$ and $\mathcal{A}_2$ are the first and last authors and $\mathcal{A}_{3-10}$ are the remaining human annotators.
Furthermore, we refer to the underlying preferences captured by an annotator's labels as a \textit{preference profile}.
Figure~\ref{fig:cm}
% in Appendix~\ref{app:ds}
shows the inter-annotator agreement of the annotators on the 200 lines (henceforth, referred to as {\dataset} \textit{(small)} dataset).
The agreement score between two users denotes the percentage of lines they both found inspiring/not inspiring.
We note that the agreement values show high fluctuations, ranging between $47.5\% - 72.5\%$, with a mean (std) agreement of $65.3 \pm 3.1\%$.
In other words, the annotators disagree on $\sim$70/200 lines on average.
Such disagreement is expected in the perception of aesthetic stimuli, as shown by~\citet{Gold9397}.
This showcases that our annotations capture diverse preference profiles of what lines different individuals consider inspiring.
Table~\ref{tab:dataset} in Appendix~\ref{app:ds} summarizes dataset statistics.

\section{Experiments}
\label{sec:method}
% \subsection{Training}
\paragraph{Step 1: Computing Features.} To test our framework on the {\dataset} dataset, we first compute all the features described in Section~\ref{sec:features}.
Specifically, we combine all the features \textit{along the time dimension} to obtain the following overall feature vector for a given lyric line:
    \begin{equation}
        \mathbf{S_{all}} = [\mathbf{S_{imagery}} | \cdots | \mathbf{S_{banality}}]
        \label{eq:s_all}
    \end{equation}
    where $|$ denotes the concatenation operation along the time dimension, $T$ is the number of time steps (= number of words in the sentence), and $\mathbf{S_{all}} \in \mathbb{R}^{T \times n}$ with $n = (8T + m)$ when using $\mathbf{S_{vfull}}$ and $n = (8T+1)$ when using $\mathbf{S_{vbin}}$.

\paragraph{Step 2: Training the Calbration Network} We create an 80-20 split of training-testing in the dataset and use the features in the training set to train the calibration network.
We use stratified sampling when creating the splits to ensure that the distribution of labels is preserved across the two dataset parts.
We perform 5-fold cross-validation for tuning the hyperparameters of the calibration network.
When training calibration network on the {\dataset} \textit{(small)} dataset, we create a 75-25 split of training-testing and then do a 5-fold cross-validation.
% \noindent\textbf{Feature Extraction.}
% Unless otherwise stated, the classifiers use the features mentioned in Section~\ref{sec:features} as input and the user-annotated labels as output in the dataset as output.

\paragraph{Choice of calibration network.}
We explore multiple deep-learning (DL) and machine-learning (ML) based architectures for the calibration network.
Namely, we try \textbf{a) LSTM+Attn} ($\mathbf{S_{all}}$): a bi-directional classification network with Long Short Term Memory (LSTM) cells~\citep{hochreiter1997long} coupled with an attention mechanism~\citep{bahdanau2014neural}. We use an LSTM network as all of our features are temporal, and LSTMs are adept at modeling long-range temporal dependencies.
% (refer to Table~\ref{tab:features} for dimensions of individual features).
% The classifier is trained on these combined features to predict whether a given lyric line is inspiring.
We also perform ablation studies where we train the classifier on a subset of $\mathbf{S_{all}}$ to test the effect of the individual features; \textbf{b) XGBoost}~\citep{chen2016xgboost}:
We train a standard XGBoost model on our features; however, we \textit{flatten} $\mathbf{S_{all}}$ as XGBoost does not inherently handle sequential data with a timestep dimension. Specifically, we train XGBoost (and all the other ML-based classifiers) on the following flattened input feature:
    \begin{equation}
        \mathbf{{S}_{flat}} = [{\mathbf{\dot{S}_{imagery}}} \oplus \cdots \oplus {\mathbf{\dot{S}_{banality}}}],
    \end{equation}
where $\mathbf{\dot{S}_{(\cdot)}}$ is the flattened version of $\mathbf{{S}_{(\cdot)}}$ and $\oplus$ is the concatenation operation.
    $\mathbf{{S}_{flat}} \in \mathbb{R}^{Tn}$, where $n=8T+m$;
\textbf{c) Other ML-based Classifiers.} In addition to XGBoost, we train multiple other ML-based classifiers on $\mathbf{{S}_{flat}}$. Specifically, we train Logistic Regression \textbf{(LR)}, Support Vector Machine \textbf{(SVM)}, Decision Trees \textbf{(DT)}, and Random Forest \textbf{(RF)} models~\citep{nelder1972generalized,cortes1995svm,quinlan1986decisiontrees,breiman2001random}.

\paragraph{Other Frameworks.}
We compare our framework with multiple strong baselines.
Specifically, we compare the following: \textbf{a) LSTM+Attn (SBERT):} we train an LSTM+Attn classifier that maps the Sentence-BERT~\citep{reimers2019sentence} embeddings of lyric lines to the final rating.
SBERT is a popular text embedding model that modifies the BERT architecture with siamese~\citep{bromley1993siamese} and triplet networks~\citep{schroff2015facenet} to obtain semantically meaningful sentence representations.
Note that in this scenario, there will be only one ``time step;'' \textbf{b) BERT}~\citep{devlin-etal-2019-bert}: a state-of-the-art transformer-based language model pre-trained on large textual corpora.
It is widely adopted for various natural language processing tasks, such as classification, question answering, and named entity recognition.
We fine-tune a Bert-base model, which has 110M parameters, on the {\dataset} dataset, which serves as a strong baseline comparing the performance of mapping the sentence embeddings of lyric lines directly with the final prediction;
    % We fine-tune a BERT classifier on the complete \textbf{EvocativeLines} dataset.
    % We do not report BERT results on \textbf{EvocativeLines (small)} as our models did not converge on the smaller version of the dataset.
\textbf{c) LLaMA-3:} a prompting-based $k$-shot LLaMA classifier, where, for a given test line, we prompt a LLaMA-3-70b model with $k$ ``inspiring'' and $k$ ``not inspiring'' examples from the dataset and ask it to classify a test sentence.
We use cosine similarity to find the $k$ examples that are (semantically) closest to the test sentence in the SBERT embedding space.
We find that this approach of selecting examples performs better than random sampling.

\paragraph{Evaluation of Different Frameworks.}
We adapt each framework on the {\dataset} dataset and 
use test accuracy and AUC scores as primary indicators of the predictive power of different models.
We report these results in Table~\ref{tab:sup_vs_semisup}.
Further, we test our proposed framework on the {\dataset} \textit{(small)} dataset with 10 diverse sets of preference profiles, and report results in Table~
\ref{tab:user_study}.
% We choose LSTM+attn as it achieves the best results amongst DL-based classifiers and XGBoost as it achieves the best results amongst ML-based classifiers.

\paragraph{Implementation Details}
\label{app:impl}
We implement the LSTM classifiers in PyTorch~\citep{paszke2019pytorch} and train them for 500 epochs with early stopping (we stop a training run if the validation performance does not improve for 10 consecutive epochs).
We perform grid search to find optimal values over the parameters listed in Table~\ref{tab:lstm_grid_search} (Appendix~\ref{app:hp_search}), and find a \textit{one}-layer \textit{bi}-directional LSTM classifier trained with a learning rate of $0.0001$ on $\mathbf{S_{all}}$ input features performs best.

We use the \texttt{sentence-transformers}\footnote{\url{https://www.sbert.net/}} python package to obtain SBERT embeddings of sentences and use Huggingface's \texttt{transformers}~\footnote{\url{https://huggingface.co/docs/transformers/index}} library to implement the fine-tuning of BERT classifier.
Specifically, we finetune the BERT classifier for 500 steps with a weight decay of $0.01$.
We use the \texttt{sklearn} python package~\citep{pedregosa2011scikit} to implement all the ML-based classifiers.
We use the default hyperparameters for all the models.
Finally, we use a batch size of 32 for all the models.

For the LLaMA-3 classifier, we use the categorization of ``inspiring'' and ``not inspiring'' from the annotations from $\mathcal{A}_1$ and $\mathcal{A}_2$ (we randomly select which annotator to choose for every test sentence).

Finally, we perform 5-fold cross-validation (except for the LLaMA-3 classifier), repeat all our experiments for 5 random seeds, and report the mean and std values in our result tables.

\begin{table}[t]
    % \small
    \centering
    \resizebox{\linewidth}{!}
    {%
    \begin{tabular}{lcccc}
        \toprule
         \textbf{Method} & \multicolumn{2}{c}{$\mathcal{A}_1$} & \multicolumn{2}{c}{$\mathcal{A}_2$} \\
         \cmidrule{2-5}
         & \textbf{Accuracy} & \textbf{AUC} & \textbf{Accuracy} & \textbf{AUC} \\
        \midrule
        \textit{Majority baseline} & \textit{51.5} & \textit{50.0} & \textit{53.0} & \textit{50.0} \\
        \midrule
        {LSTM+Attn} ($\mathbf{S_{all}}$) & \mustdbf{79.5}{0.4} & \mustdbf{79.7}{0.1} & \mustdbf{78.3}{0.3} & \mustdbf{80.1}{0.3} \\
        {LSTM+Attn (SBERT)} & \mustd{71.6}{0.3} & \mustd{71.7}{0.1} & \mustd{69.9}{0.2} & \mustd{70.1}{0.3} \\
        BERT & \mustd{77.6}{2.1} & \mustd{78.8}{1.7} & \mustd{75.2}{2.3} & \mustd{76.1}{1.1} \\
        \midrule
        RF & \mustd{80.5}{1.7} & \mustd{80.6}{1.7} & \mustd{72.4}{2.0} & \mustd{73.0}{1.2} \\
        SVM & \mustd{70.5}{1.2} & \mustd{70.5}{1.0} & \mustd{67.3}{2.3} & \mustd{63.1}{1.1} \\
        LR & \mustd{71.1}{1.9} & \mustd{71.1}{1.8} & \mustd{64.3}{1.9} & \mustd{62.0}{1.4} \\
        DT & \mustd{84.2}{2.5} & \mustd{84.3}{2.3} & \mustd{84.1}{1.0} & \mustd{80.0}{1.7} \\
        XGBoost & \mustdbf{92.2}{1.7} & \mustdbf{92.2}{1.6} & \mustdbf{87.32}{1.4} & \mustdbf{83.9}{1.2} \\
        \midrule
        LLaMA (200-shot) & \mustd{73.3}{1.7} & \mustd{73.3}{1.6} & \mustd{62.3}{3.6} & \mustd{65.1}{1.6} \\
        LLaMA (300-shot) & \mustd{73.9}{2.1} & \mustd{74.2}{1.8} & \mustd{64.1}{1.2} & \mustd{65.8}{1.4} \\
        LLaMA (450-shot) & \mustdbf{74.8}{3.1} & \mustdbf{74.8}{3.0} & \mustdbf{69.5}{3.5} & \mustdbf{70.5}{1.6} \\
        \bottomrule
    \end{tabular}
    }
    \caption{Comparison of different frameworks on {\dataset}. The first section shows the performance of DL-based methods, the second section shows the performance of ML-based methods, and the last section shows the performance of LLM-based methods.}
    \label{tab:sup_vs_semisup}
\end{table}

\begin{table}[t]
\centering
\resizebox{\linewidth}{!}{
    \begin{tabular}{lcccccc}
    \toprule
     User & \multicolumn{2}{c}{\textbf{Majority}} & \multicolumn{2}{c}{\textbf{LSTM+Attn}} & \multicolumn{2}{c}{\textbf{XGBoost}} \\
     \cmidrule{2-7}
     & \textbf{Accuracy} & \textbf{AUC} & \textbf{Accuracy} & \textbf{AUC} & \textbf{Accuracy} & \textbf{AUC} \\
    \midrule
    \(\mathcal{A}_1\) & 50.0 & 50.0 & \mustd{75.6}{1.2} & \mustd{64.9}{4.0} & \mustd{70.7}{5.3} & \mustd{70.2}{5.3} \\
    \(\mathcal{A}_2\) & 70.5 & 50.0 & \mustd{80.5}{1.2} & \mustd{62.3}{6.1} & \mustd{69.3}{9.5} & \mustd{57.1}{8.4} \\
    \(\mathcal{A}_3\) & 58.5 & 50.0 & \mustdbf{94.1}{1.2} & \mustdbf{92.3}{2.1} & \mustd{63.4}{11.1} & \mustd{62.6}{11.7} \\
    \(\mathcal{A}_4\) & 60.5 & 50.0 & \mustd{68.7}{3.1} & \mustd{61.4}{4.5} & \mustd{59.5}{4.5} & \mustd{56.4}{6.6} \\
    \(\mathcal{A}_5\) & 58.0 & 50.0 & \mustd{80.4}{2.5} & \mustd{84.8}{5.1} & \mustd{71.7}{6.7} & \mustdbf{70.5}{6.6} \\
    \(\mathcal{A}_6\) & 64.5 & 50.0 & \mustd{75.6}{3.1} & \mustd{75.4}{2.8} & \mustd{68.8}{3.2} & \mustd{62.8}{3.0} \\
    \(\mathcal{A}_7\) & 58.0 & 50.0 & \mustd{75.6}{4.3} & \mustd{82.3}{4.2} & \mustd{61.0}{4.4} & \mustd{60.3}{4.2} \\
    \(\mathcal{A}_8\) & 62.5 & 50.0 & \mustd{80.5}{5.4} & \mustd{78.2}{5.1} & \mustdbf{72.2}{5.0} & \mustd{68.5}{7.1} \\
    \(\mathcal{A}_9\) & 70.0 & 50.0 & \mustd{76.5}{4.7} & \mustd{72.4}{3.6} & \mustd{70.7}{4.9} & \mustd{57.3}{5.1} \\
    \(\mathcal{A}_{10}\) & 58.5 & 50.0 & \mustd{82.3}{2.3}& \mustd{84.2}{3.5} & \mustd{62.4}{5.9} & \mustd{60.7}{5.0} \\
    \midrule
    \textbf{Avg.} & 61.1 & 50.0 & \mustdbf{79.0}{6.3}& \mustdbf{75.8}{9.9} & \mustd{67.0}{4.6} & \mustd{62.6}{5.1} \\
    \bottomrule
    \end{tabular}
}
\caption{Performance of the proposed framework with LSTM+Attn and XGBoost calibration networks on the \textit{EvocativeLines (small)} dataset.}
\label{tab:user_study}
\end{table}

% \begin{table*}[htbp!]
%     % \small
%     \centering
%     \resizebox{\linewidth}{!}{
%     \begin{tabular}{ccccccccccc}
%         \toprule
%          \textbf{Accuracy} ($\mathcal{A}_1$) & \textbf{Accuracy} ($\mathcal{A}_2$) & $\mathbf{imagery}$ & $\mathbf{energy}$ & $\mathbf{abs}$ & $\mathbf{vfull}$ & $\mathbf{npmi_{uni}}$ & $\mathbf{npmi_{bi}}$ & $\mathbf{surprisal}$ & $\mathbf{entropy}$ & $\mathbf{banality}$ \\
%         \midrule
%         & & \ding{55} & & & & & & & &  \\
%         & & & \ding{55} & & & & & & &  \\
%         & & & & \ding{55}  & & & & & &  \\
%         & & & & & \ding{55} & & & & &  \\
%         & & & & & & \ding{55} & & & &  \\
%         & & & & & & & \ding{55} & & &  \\
%         & & & & & & & & \ding{55} & &  \\
%         & & & & & & & & & \ding{55} &  \\
%         & & & & & & & & & & \ding{55}  \\
%         & & \ding{55} & \ding{55} & \ding{55} & \ding{55} & & & & &  \\
%         & & & & & & \ding{55} & \ding{55} & \ding{55} & \ding{55} & \ding{55} \\
%         \bottomrule
%     \end{tabular}
%     }
%     \caption{Ablation on the proposed features on {\dataset}. \ding{55} denotes removal of the feature.}
%     \label{tab:ablation}
% \end{table*}

\section{Quantitative Results}
\label{sec:results}
% Table~\ref{tab:sup_vs_semisup} and Table~\ref{tab:user_study} present the results of our experiments on {\dataset} and {\dataset} \textit{(small)}, respectively.
% We present our findings below.

\subsection{Proposed Linguistic and Poetic Features Have a High Predictive Power}
From Table~\ref{tab:sup_vs_semisup}, we note that our proposed framework with ML-based and DL-based calibration networks outperforms a 450-shot LLaMA-3-70b classifier.
Specifically, our XGBoost model with just 163 parameters achieved a test accuracy of 92.2 for \aid{1} and 87.3 for \aid{2}, compared to 450-shot LLaMA-3 classifier with 70B parameters that achieves an accuracy of 74.8 and 69.8 for \aid{1} and \aid{2}, respectively.
Note that 450 was the maximum number of positive and negative examples we were able to add to the prompt of the LLaMA-3-70b classifier.
Among the DL-based models, the LSTM+Attn model (with roughly 110k parameters) outperforms a BERT model (with 110M params) fine-tuned on the {\dataset} dataset.
It also outperforms the LSTM+Attn (SBERT) baseline, showcasing the higher predictive power of the proposed features.

These higher predictive power of our features computed using an LLM compared to using the LLM out-of-the-box shows that while we can leverage the knowledge of LLMs to compute well-established linguistic and stylistic characteristics of a lyric line, the LLM itself has difficulty in capturing the nuanced differences between what is perceived as ``inspiring'' and ``not inspiring'' by an individual.
Overall, this highlights that our proposed framework that identifies the key linguistic and poetic features is more suitable for modeling artistic preferences of lyrical lines compared to using significantly larger models out-of-the-box.

\begin{table*}[ht]
\centering
\resizebox{0.8\linewidth}{!}
{%
\begin{tabular}{lcccccccccc}
\toprule
\textbf{Feature} & \multicolumn{2}{c}{$\mathcal{A}_1$} & \multicolumn{2}{c}{$\mathcal{A}_2$} & \multicolumn{2}{c}{$\mathcal{A}_3$} & \multicolumn{2}{c}{$\mathcal{A}_6$} & \multicolumn{2}{c}{$\mathcal{A}_7$} \\
\cmidrule{2-11}
{} & coef & $p$-value & coef & $p$-value & coef & $p$-value & coef & $p$-value & coef & $p$-value \\
\midrule
% \textit{const} & -5.6493 & 0.375 & -2.4286 & 0.114 & -3.8495 & 0.01 & -2.0376 & 0.15 & 1.9059 & 0.188 \\
$\mathbf{\overline{S}_{surprisal}}$ & -0.4149 & 0.107 & 0.288 & 0.404 & 0.4351 & 0.233 & 0.416 & 0.242 & -0.2176 & 0.518 \\
$\mathbf{\overline{S}_{entropy}}$ & -0.8377 & \highlight{0.034} & 0.0534 & 0.889 & 0.02 & 0.958 & -0.0632 & 0.873 & -0.0817 & 0.822
 \\
$\mathbf{\overline{S}_{npmi,uni}}$ & 7.2427 & \highlight{0.000} & 9.7483 & \highlight{0.068} & 59.1835 & \highlight{0.003} & 75.9894 & \highlight{0.018} & 21.0617 & 0.152
 \\
$\mathbf{\overline{S}_{npmi,bi}}$ & -6.2588 & \highlight{0.002} & -9.8577 & 0.552 & -55.0718 & \highlight{0.007} & -73.9488 & \highlight{0.024} & -19.2738 & 0.224
 \\
$\mathbf{\overline{S}_{abs}}$ & 0.5402 & 0.298 & 0.6791 & \highlight{0.047} & 0.7423 & \highlight{0.045} & 0.9499 & \highlight{0.012} & 0.7226 & \highlight{0.03}
 \\
$\mathbf{\overline{S}_{vbin}}$ & 0.3281 & \highlight{0.001} & -103.4697 & 0.556 & -197.0881 & \highlight{0.028} & -264.4955 & 0.158 & 9.1933 & 0.957
 \\
$\mathbf{\overline{S}_{imagery}}$ & 14.8346 & \highlight{0.000} & 0.6861 & \highlight{0.048} & 2.065 & 0.057 & 0.4584 & 0.643 & 1.1321 & 0.241
 \\
$\mathbf{\overline{S}_{energy}}$ & -0.0543 & 0.985 & 0.9405 & \highlight{0.031} & 1.3163 & 0.423 & 2.4855 & 0.086 & 1.8852 & 0.191
 \\
$\mathbf{\overline{S}_{banality}}$ & -2.8474 & 0.128 & 0.5269 & 0.636 & -0.2816 & 0.804 & 0.1945 & 0.866 & 3.2007 & \highlight{0.003}
 \\
 \midrule
$\mathbf{\overline{S}_{I1}}$ & 0.0549 & 0.084 & -0.1213 & 0.609 & -0.0655 & 0.745 & -0.2414 & 0.255 & 0.3546 & 0.116
 \\
$\mathbf{\overline{S}_{I2}}$ & -0.0188 & 0.499 & 13.8702 & \highlight{0.027} & 14.5632 & 0.142 & 6.5436 & 0.438 & 3.1841 & 0.72 \\
$\mathbf{\overline{S}_{all}}$ & 0.0760 & \highlight{0.025} & -1.0131 & \highlight{0.022} & -0.3216 & 0.31 & -0.2537 & 0.306 & 1.3457 & 0.145 \\
\bottomrule
\end{tabular}
}
\caption{Results of interaction testing. I1 denotes the interaction between linguistic features (surprisal, entropy, banality, npmi), and I2 denotes the interaction between stylistic features (imagery, word energy, abstraction, valence).}
\label{tab:interaction}
\end{table*}

\subsection{The Proposed Framework is Robust to Diverse User Preferences}
Table~\ref{tab:user_study} shows the results for our experiments on {\dataset} \textit{(small)} that contains 10 preference profiles captured by the 10 annotators.
Specifically, we note that our framework with LSTM+Attn and XGBoost calibration network can adapt to the diverse range of artistic preferences captured in the {\dataset} \textit{(small)} dataset, with the LSTM+Attn variant achieving a test accuracy of 79.0 $\pm$ 6.3\% average across the 10 annotators and the XGBoost variant achieving 67 $\pm$ 4.6\% average test accuracy.
% We choose the LSTM+Attn and XGBoost models as they achieve the best performance among all the DL- and ML-based baselines on the {\dataset} dataset with 3,025 annotations.
We notice a significant drop in the performance of XGBoost relative to the performance on the full dataset,
% It achieves an average test accuracy and AUC of 67.0\% and 62.6\% across the 10 annotators.
which can be explained by the fact that while XGBoost is a flexible model, it might suffer from overfitting as we shift to data-scarce setups.
% On the other hand, the LSTM+Attn classifier maintains its decent level of performance and achieves an average test accuracy and AUC of 79.0\% and 75.8\%, respectively, across all the annotators.
% We only report results for our framework  it significantly outperforms the other frameworks.
Overall, these results highlight the robustness of our framework against the high degree of variations across different preference profiles, and we conclude that the proposed framework can be successfully employed to forecast the artistic preferences of different individuals, even in low-data setups, by choosing an appropriate calibration network.

\subsection{Significance of Individual Features}
% Prior subsections demonstrate the efficacy of our framework in capturing diverse user preferences.
To understand the feature interactions at a more fine-grained level and how they control the final outcome, we conduct statistical testing to determine the correlations amongst the individual features and their influence on the output variable (whether or not an individual found a lyric line inspiring).
Specifically, we fit a logistic regression model with interaction terms to evaluate the significance of each independent variable (feature) on the binary outcome variable~\footnote{we use \texttt{statsmodels} python package to implement logistic regression with interaction terms.}.
We simplify our features for this test by taking a mean across the timesteps dimension so that each feature is converted into a score.
While this simplified representation is not truly representative of our actual features, it serves as a good proxy.
We show the results of significant testing in Table~\ref{tab:interaction} for \aid{1}, \aid{2}, \aid{3}, \aid{6}, and \aid{7}.
For brevity, we only show results for five annotators.
We chose \aid{1} and \aid{2} as their annotations allow us to study interactions in a data-rich setting, and we chose \aid{3}, \aid{6}, and \aid{7} as this triplet has one of the lowest inter-annotator agreement in the smaller version of our dataset (refer to Figure~\ref{fig:cm}).
We also study three composite interactions: \textbf{a)} linguistic interaction between surprisal, entropy, banality, and npmi (uni and bi) (denoted by $\mathbf{\overline{S}_{I1}}$); \textbf{b)} poetic and artistic interaction between imagery, word energy, level of abstraction, and valence (denoted by $\mathbf{\overline{S}_{I2}}$); and \textbf{c)} interaction among all the features (denoted by $\mathbf{\overline{S}_{all}}$).

We report the results of our interaction study in Table~\ref{tab:interaction}.
Starting with \aid{1}, we note that $\mathbf{\overline{S}_{entropy}}$ has a low $p$-value of 0.034 (< 0.05) and a negative coefficient of -0.8377.
Therefore, we conclude that contextual entropy has a \textit{significantly negative} correlation with the outcome variable.
In other words, as the entropy of a lyric line increases, \aid{1} is likely to find that lyric line evocative.
We follow a similar analysis pattern to conclude that \s{npmi,uni} is \textit{significantly positively} correlated and \s{npmi,bi} is \textit{significantly negatively} correlated with the outcome.
Simultaneously, poetic imagery has a \textit{significantly positive} correlation with a high coefficient of 14.8346.
This contradicting preference in word associations based on directionality coupled with a high preference for imagery indicates a strong inclination for causality in lyrics, where the line has a clear linear progression but loses its creative mark upon reversing the chain of events.
Similarly, we conclude that \s{vbin} is \textit{significantly positively} correlated with the outcome variable, indicating a preference for positive lines by \aid{1}.
Finally, we find \s{all} to be \textit{positively} correlated with the outcome variable.
We conduct a similar analysis for the other annotators and highlight the significant features for each of them in Table~\ref{tab:interaction}.
These results show that each annotator's artistic preferences were influenced by a different set of features, and our framework can be used to explain aesthetic preferences in an interpretable manner.
% \comment{talk about vbin and all, then for the others just say which features are important and what does that indicate.}

% \begin{figure*}[h!]
% \centering
% \begin{tabular}{ccc}
%     \includegraphics[width=0.3\textwidth]{figures/kde_1273.png} & \includegraphics[width=0.3\textwidth]{figures/kde_3253.png} & \includegraphics[width=0.3\textwidth]{figures/kde_7321.png} \\
%     \includegraphics[width=0.3\textwidth]{figures/kde_7326.png} & \includegraphics[width=0.3\textwidth]{figures/kde_7923.png} & \includegraphics[width=0.3\textwidth]{figures/kde_8731.png} \\
%     \includegraphics[width=0.3\textwidth]{figures/kde_8912.png} & \includegraphics[width=0.3\textwidth]{figures/kde_9361.png} & \includegraphics[width=0.3\textwidth]{figures/kde_gt.png} \\
% \end{tabular}
% \caption{KDE Plots for user distributions.}
% \end{figure*}

\section{Qualitative Examples}
We include a radar chart of different preference profiles in Figure~\ref{fig:profiles} of Appendix~\ref{app:ds} that shows how the artistic preferences of different individuals vary from one another.
We construct a preference profile by computing the average value for the positive and negative samples across the different feature dimensions we have.
We denote the profile of positive samples in blue and negative samples in red.
We notice that all the preference profiles show a visibly different distinction between the positive and negative examples.
Most prominently, all profiles have a higher imagery and energy value for the positive examples compared to the negative examples.
Additionally, we also show radar charts in Figure~\ref{fig:web_diagram_grid_4x7_inverted_banality} of Appendix~\ref{app:web_lines} that visualize some positive and negative examples based on their feature values along the different feature dimensions.
These charts show how different concepts in the lyrical content influence the different dimensions of the features.
For instance, ``all the world is the fire'' (row 1 column 2) has a 0 valence value but high imagery and energy values.

% things left:
% - discussion \\
% - dataset examples with radar charts for the lines \\
% - related work \\
% - highlight the framework part of the work. currently, methodology discusses the extract-then-train part of the work, which is more for testing/validation \\
% - have an algo for identifying lines that fit a certain knob config \\
% - add extra ablation results with the actual model too \\
% - can we design an experiment where we simulate certain artist preferences in the dataset then try to retrieve them? no \\
% - 

\section{Conclusion}
In this work, we propose an interpretable and flexible framework for computational modeling of artistic inspiration in the context of AI-generated lyrical lines that can be used to assist artists in the Seed phase of creation by identifying potentially ``inspiring'' ideas (or seeds) amongst a larger pool of candidates.
We identified the key linguistic and stylistic features that can accurately forecast aesthetic preferences among a diverse set of individuals.
Our experiments demonstrate that the predictive power of our framework surpasses that of several orders of magnitude larger state-of-the-art language models like LLaMA-3-70b.

Crucially, our approach is not limited to the analysis of lyrics and poetry and can be adopted (with a different or expanded set of features) to predict other aesthetic, literary, or artistic preferences that are inherently subjective (for instance, comparing sonnets from multiple different artists).
Overall, in the context of creating art, it is crucial to measure how well AI systems support individual creative processes that can encompass a wide spectrum of artistic preferences and skill levels, and through this work, we hope to provide a foundation for future research in this area. 
% Through this work, we aim to gain insights into the complex internal process governing subjective preferences in diverse individuals.

\bibliography{tacl2021}
\bibliographystyle{acl_natbib}

% \iftaclpubformat

% \onecolumn

\appendix

\section{Hyperparameter Selection}
\label{app:hp_search}

\begin{table}[htbp!]
    \centering
    \resizebox{\linewidth}{!}{
    \begin{tabular}{lll}
    \toprule
    \textbf{Parameter} & \textbf{Search Space} & \textbf{Optimal Value} \\
    \midrule
       input feature & $\{\mathbf{S_{all}}, \mathbf{S_{flat}}\}$ & $\mathbf{S_{all}}$ \\
        bi-directional & \{True, False\} & True \\
       attention type & \{global, general, self\} & global \\
       layers & $\{1, 2, 3\}$ & 1 \\
       hidden dimensions & $\{8, 16, 32, 64, n/2\}$ & $n$ \\
       learning rate & $10^{-i} \ \forall \ i \in \{1,\dots, 5\}$ & $10^{-4}$ \\
    \bottomrule
    \end{tabular}
    }
    \caption{Search space for grid search of LSTM. \textbf{Note:} $n$ is from Equation~\ref{eq:s_all}}.
    \label{tab:lstm_grid_search}.
\end{table}

\section{Dataset Statistics and User Profiles}
\label{app:ds}

\begin{table}
    \centering
    \resizebox{0.8\linewidth}{!}{
    \begin{tabular}{ll}
    \toprule
    \textbf{Feature Name} & \textbf{Notation} \\
    \midrule
        Poetic Imagery & $\mathbf{S_{imagery}} \in \mathbb{R}^{T \times T}$ \\
        Word Energy & $\mathbf{S_{energy}} \in \mathbb{R}^{T \times T}$ \\
        Level of Abstraction & $\mathbf{S_{abs}} \in \mathbb{R}^{T \times T}$ \\
        Valence (full) & $\mathbf{S_{vfull}} \in \mathbb{R}^{T \times m}$ \\
        Valence (binary) & $\mathbf{S_{vbin}} \in \mathbb{R}^{T}$ \\
        Surprisal & $\mathbf{S_{surprisal}} \in \mathbb{R}^{T \times T}$ \\
        Contextual Entropy & $\mathbf{S_{entropy}} \in \mathbb{R}^{T \times T}$ \\
        NPMI (uni-dir.) & $\mathbf{S_{npmi,uni}} \in \mathbb{R}^{T \times T}$ \\
        NPMI (bi-dir.) & $\mathbf{S_{npmi,bi}} \in \mathbb{R}^{T \times T}$ \\
        Banality & $\mathbf{S_{banality}} \in \mathbb{R}^{T \times T}$ \\
    \bottomrule
    \end{tabular}
    }
    \caption{Summary of features and their dimensions. Here, $T$ is the number of words (or time steps) in the input sequence, and $m$ is the total number of emotions (29, in our case).}
    \label{tab:features}
\end{table}

\begin{table}[]
    \centering
    \resizebox{0.5\linewidth}{!}{
    \begin{tabular}{lcc}
    \toprule
    \textbf{Annotator} & \#pos & \#neg \\
    \midrule
    \multicolumn{3}{c}{\textbf{EvocativeLines}} \\
    \midrule
    $\mathcal{A}_1$ & 1466 & 1559  \\
    $\mathcal{A}_2$ & 1422 & 1603 \\
    \midrule
    \multicolumn{3}{c}{\textbf{EvocativeLines (small)}} \\
    \midrule
    $\mathcal{A}_1$ & 100 & 100 \\
    $\mathcal{A}_2$ & 59 & 141 \\
    $\mathcal{A}_3$ & 117 & 83 \\
    $\mathcal{A}_4$ & 121 & 79 \\
    $\mathcal{A}_5$ & 116 & 84 \\
    $\mathcal{A}_6$ & 71 & 129 \\
    $\mathcal{A}_7$ & 112 & 88 \\
    $\mathcal{A}_8$ & 77 & 123 \\
    $\mathcal{A}_9$ & 140 & 60 \\
    $\mathcal{A}_{10}$ & 117 & 83 \\
    \bottomrule
    \end{tabular}
    }
    \caption{Dataset statistics, broken down by annotator.}
    \label{tab:dataset}
\end{table}

\begin{figure*}[h!]
\centering
\resizebox{0.9\linewidth}{!}{
\begin{tabular}{ccc}
    \includegraphics[width=0.32\textwidth]{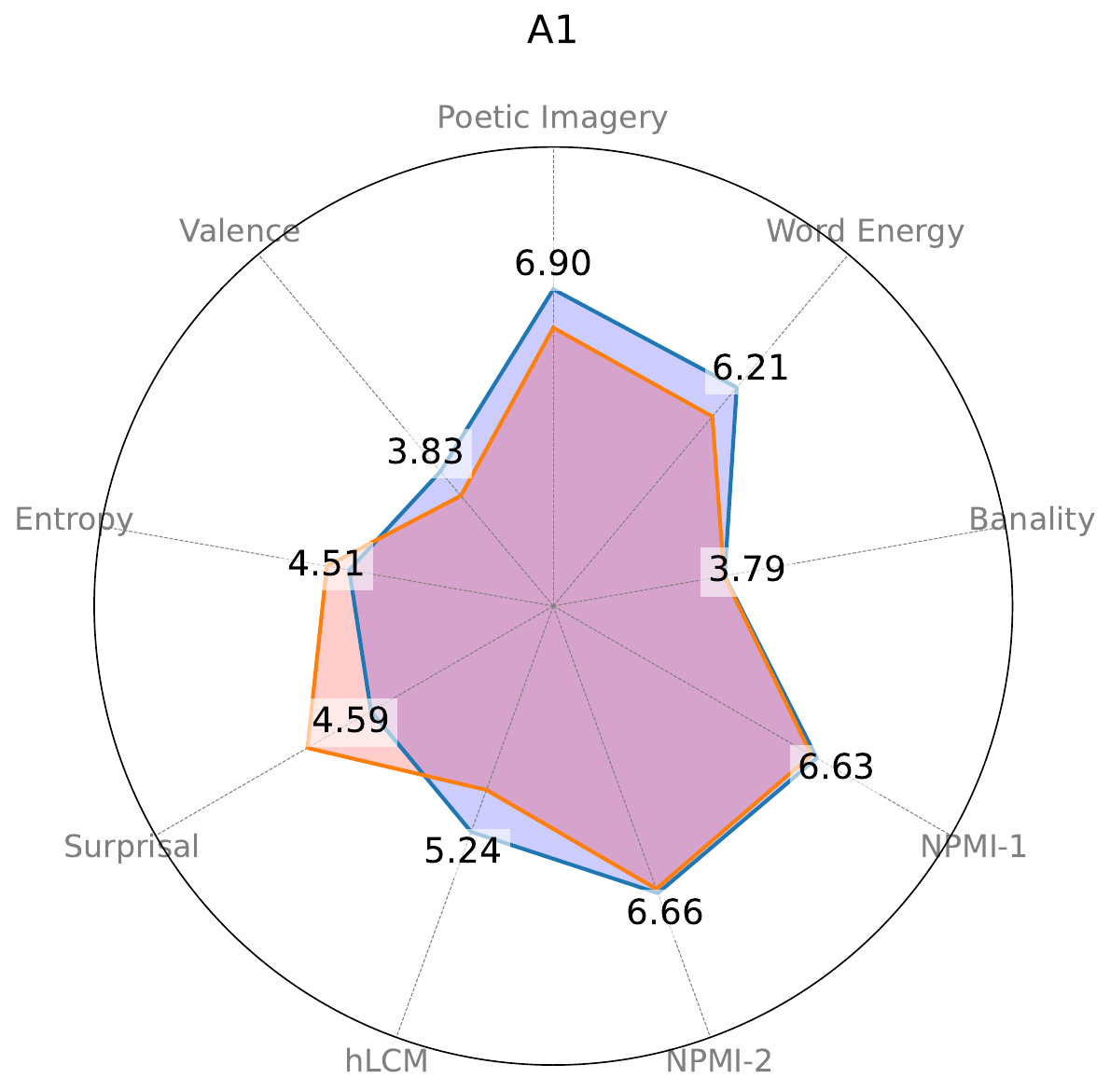} & \includegraphics[width=0.32\textwidth]{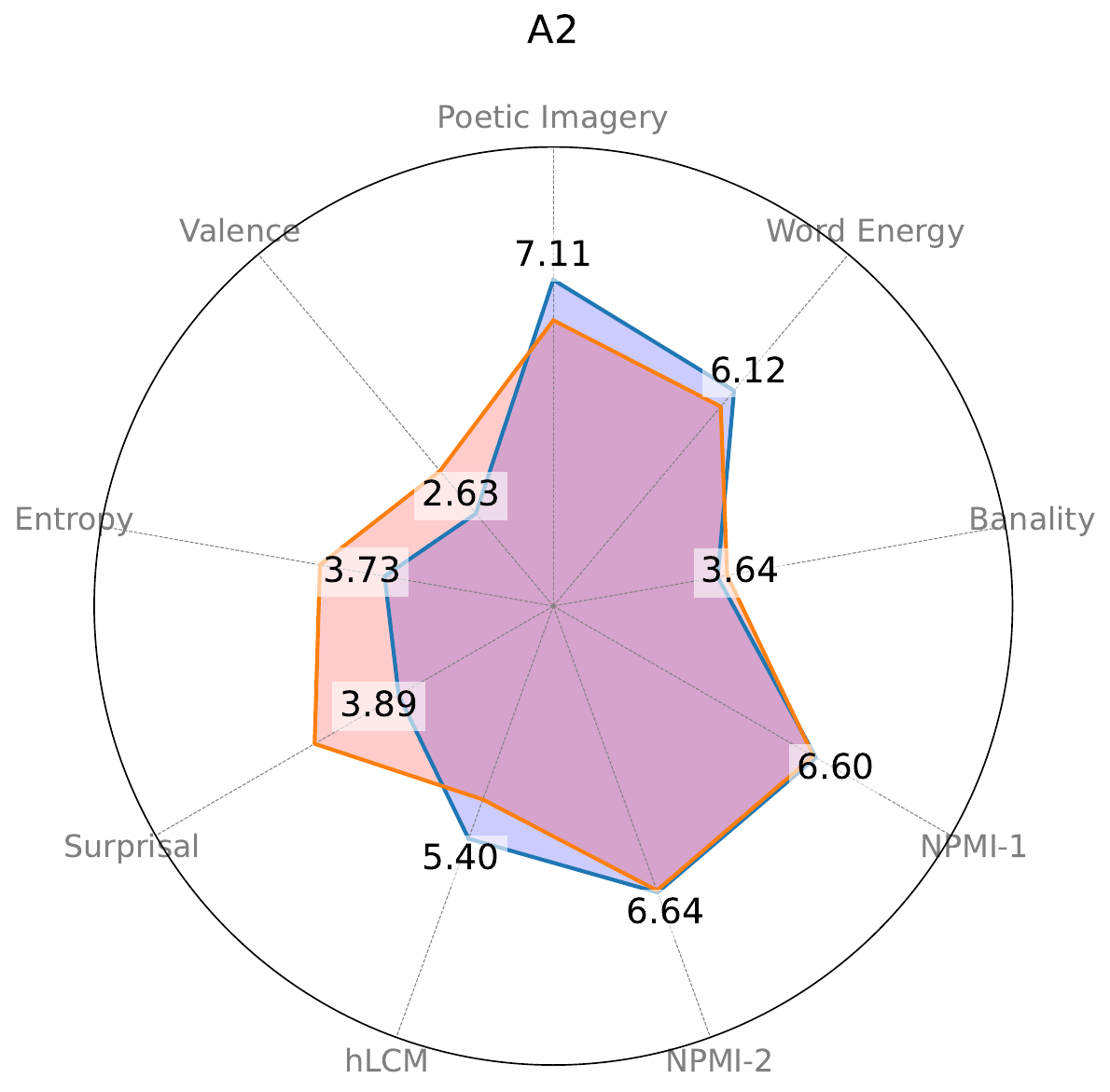} & \includegraphics[width=0.32\textwidth]{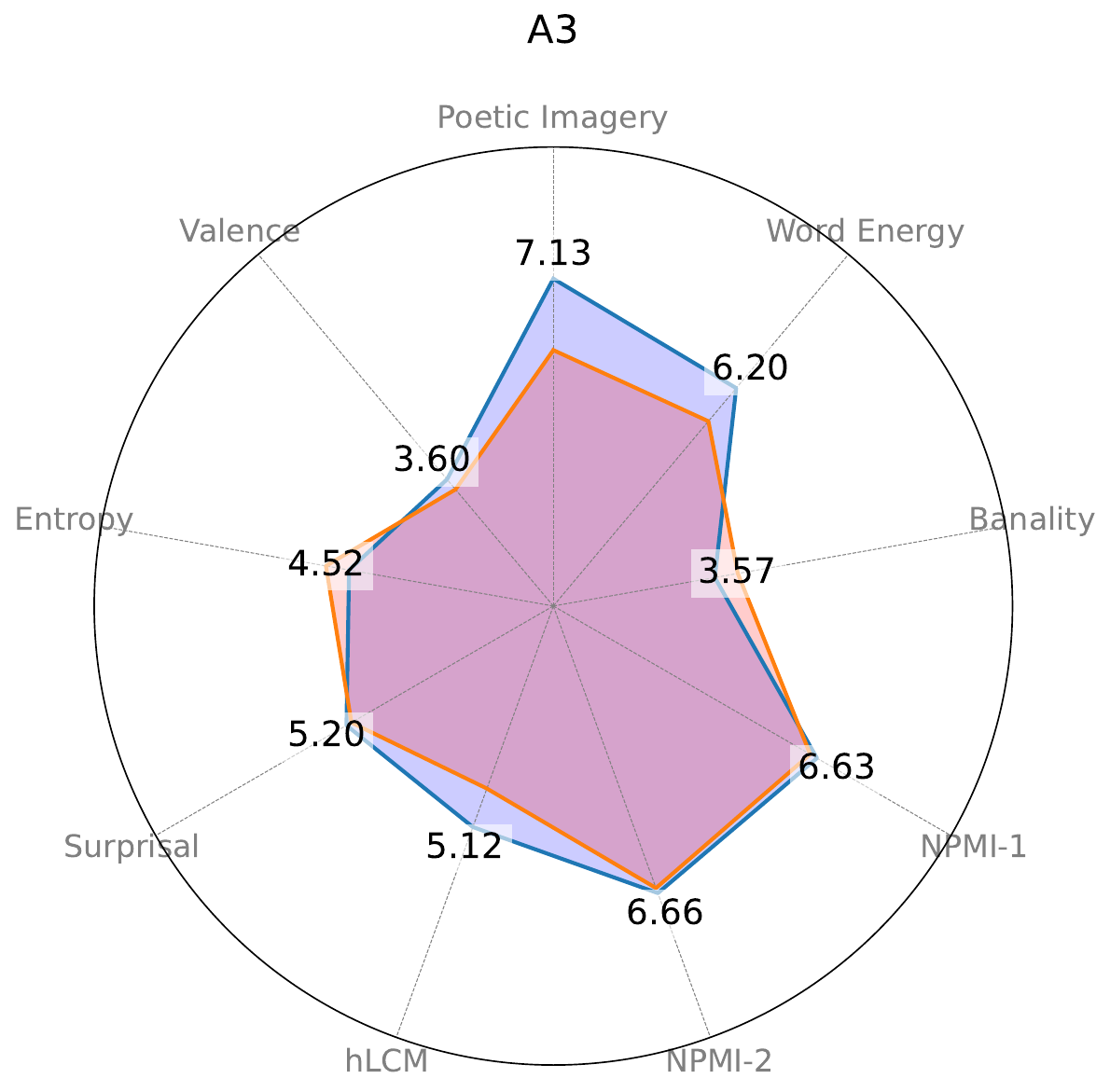} \\
\includegraphics[width=0.32\textwidth]{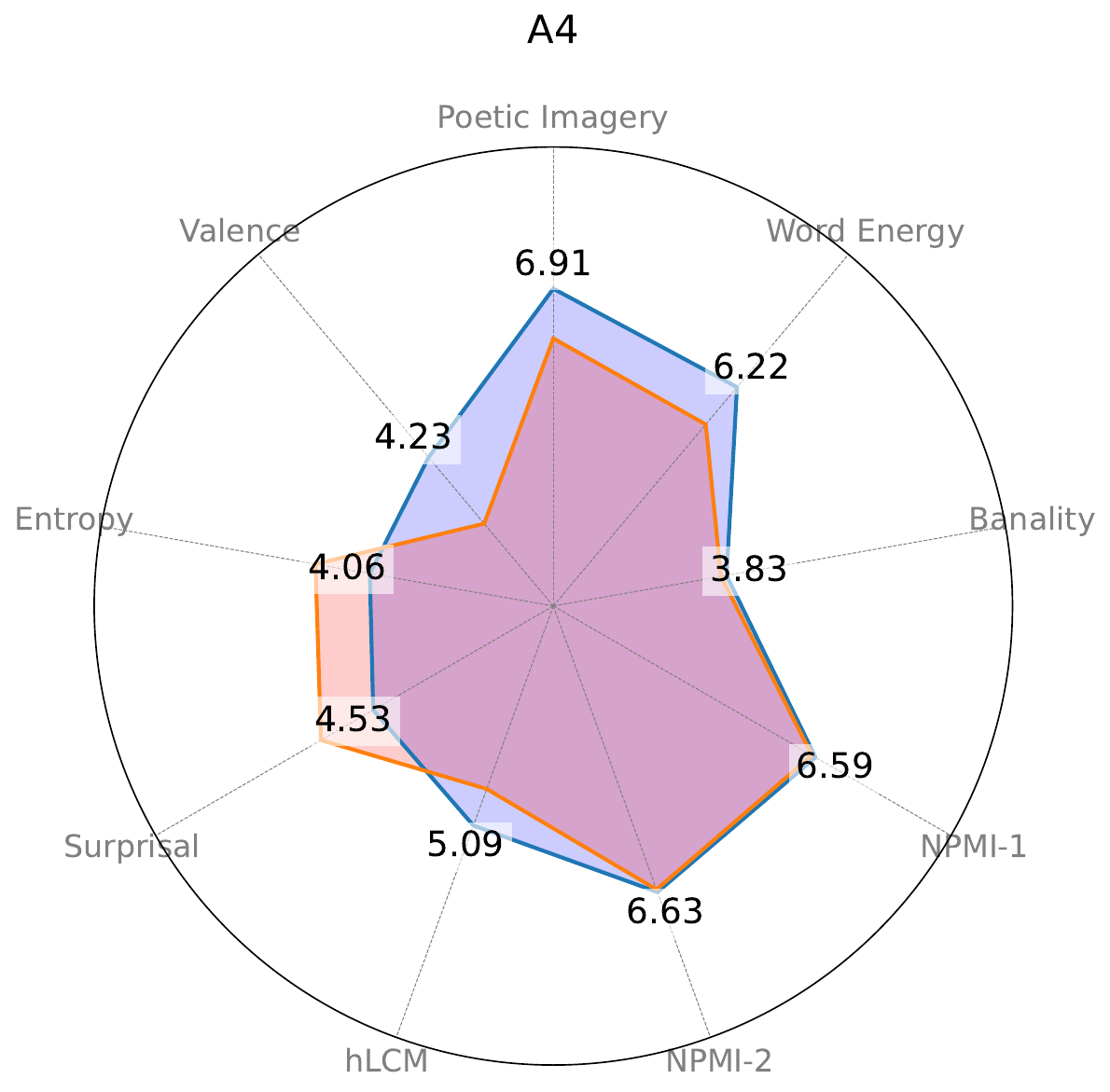} & \includegraphics[width=0.32\textwidth]{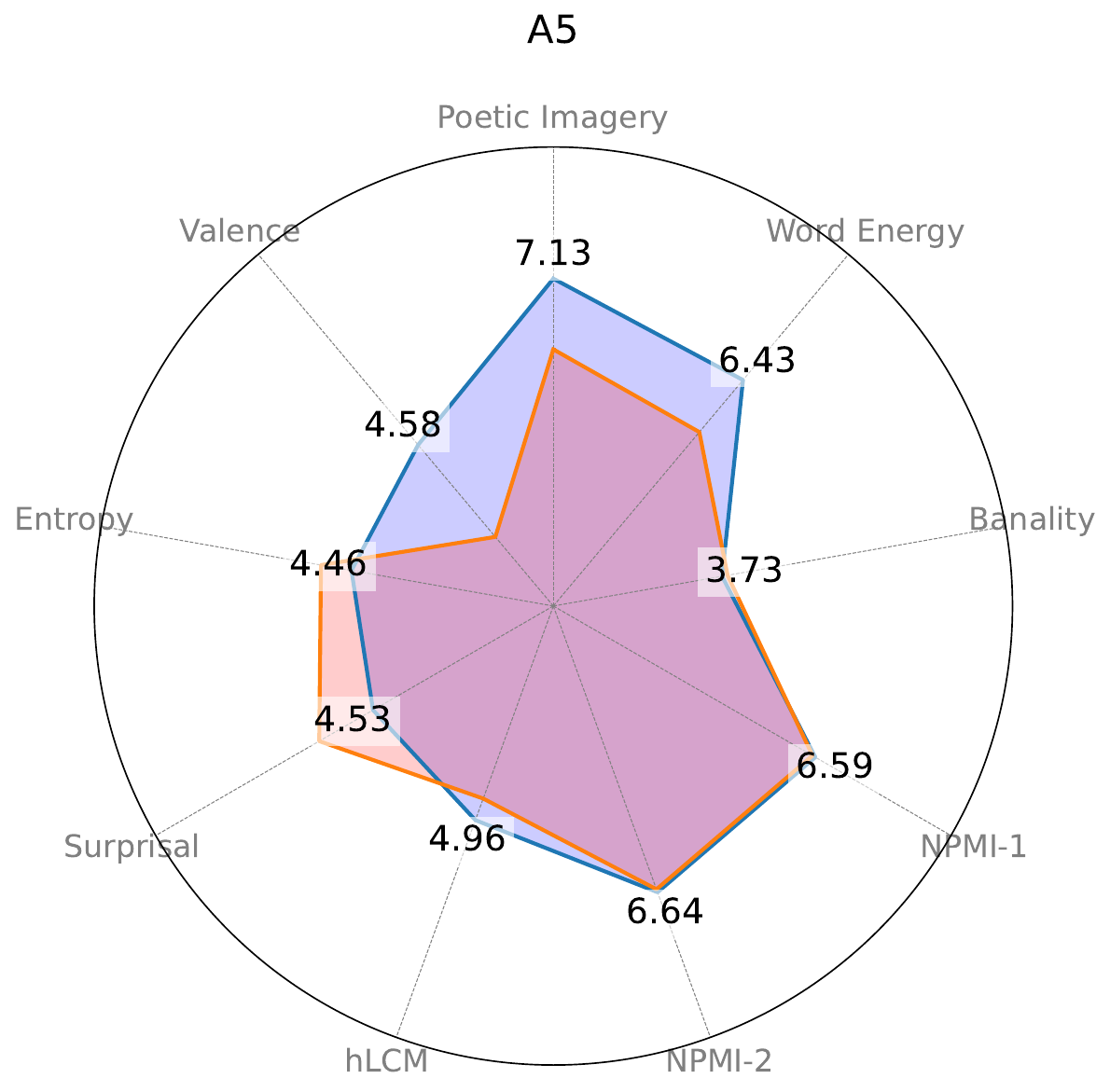} & \includegraphics[width=0.32\textwidth]{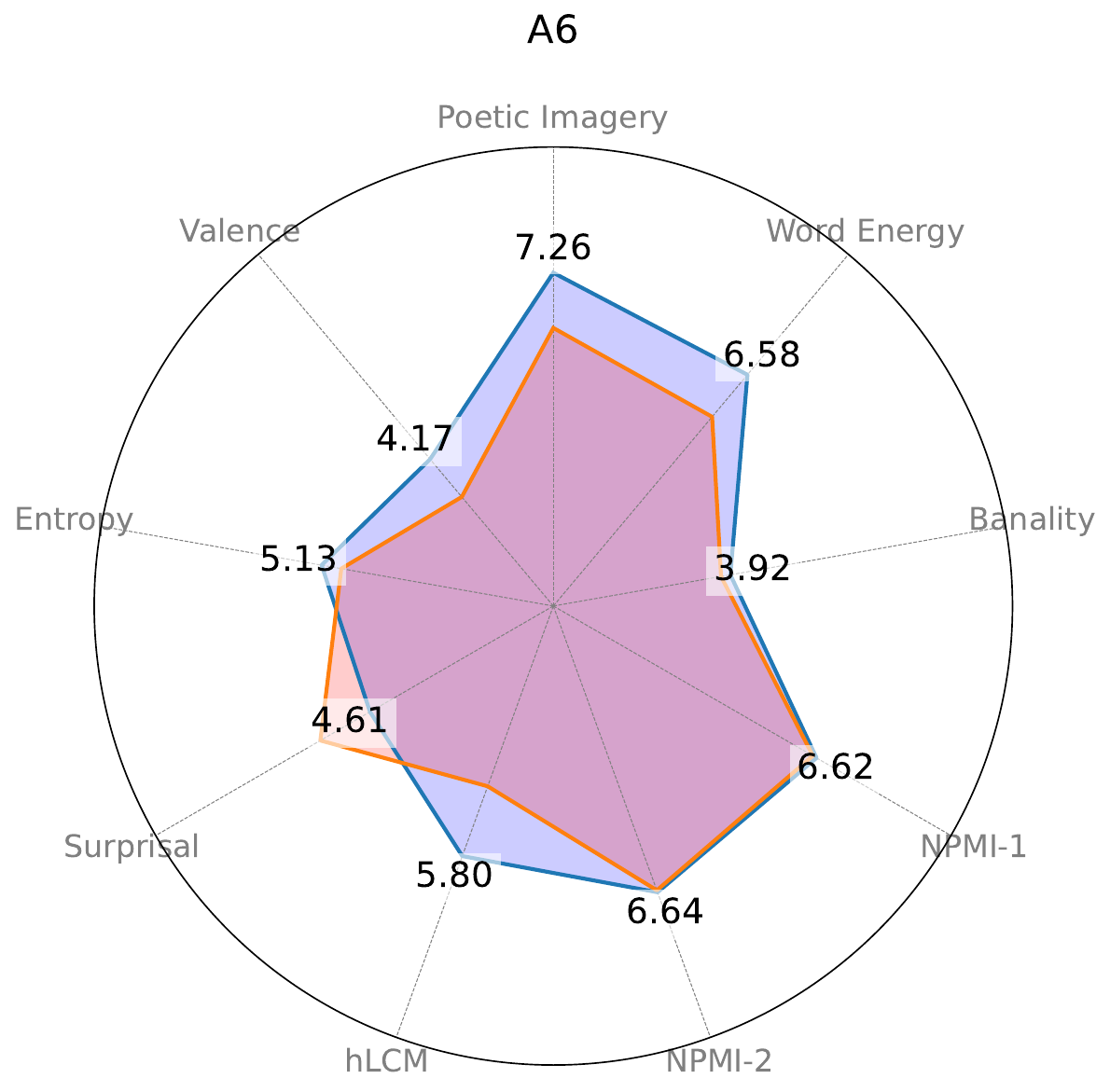} \\\includegraphics[width=0.32\textwidth]{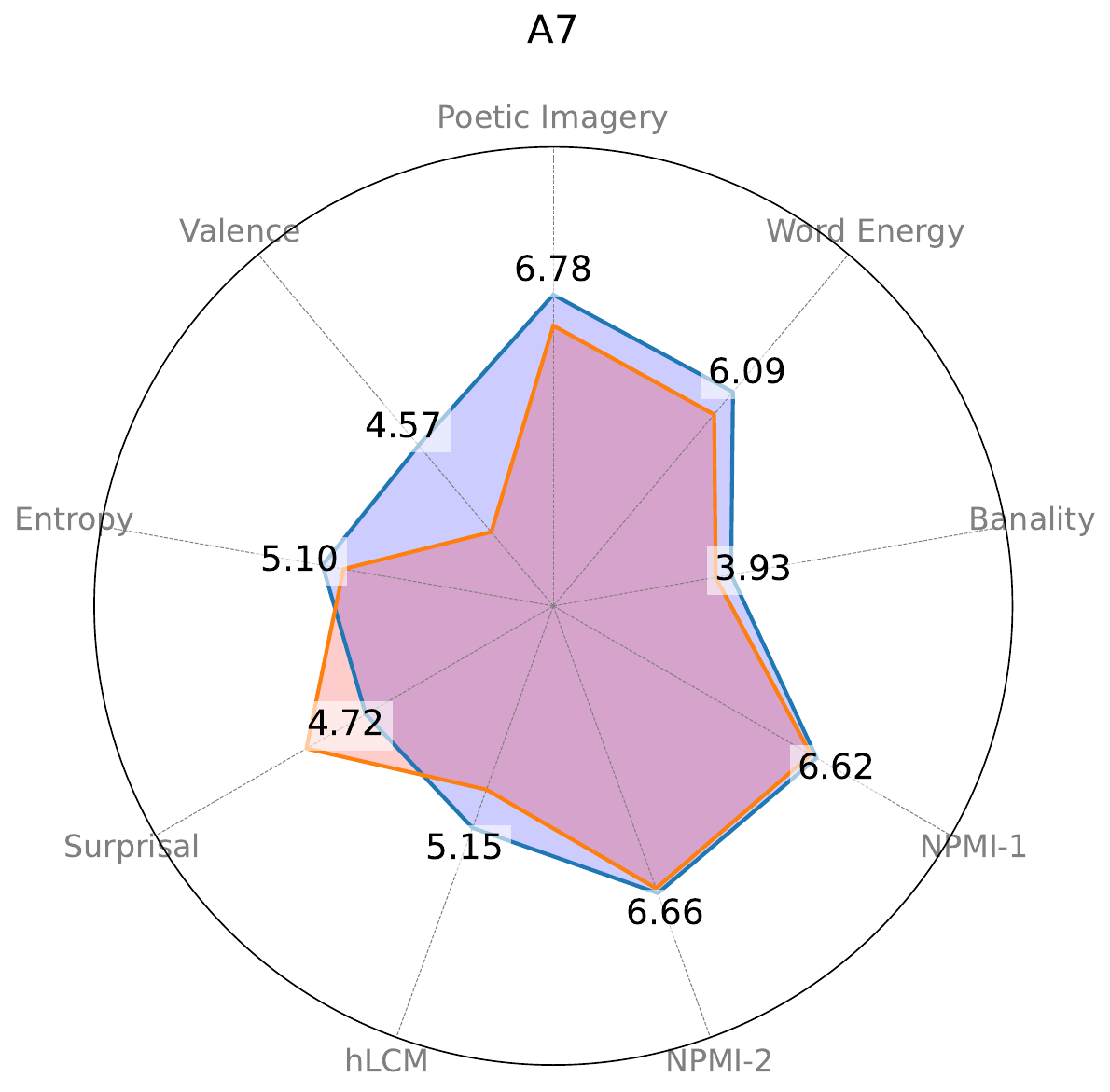} & \includegraphics[width=0.32\textwidth]{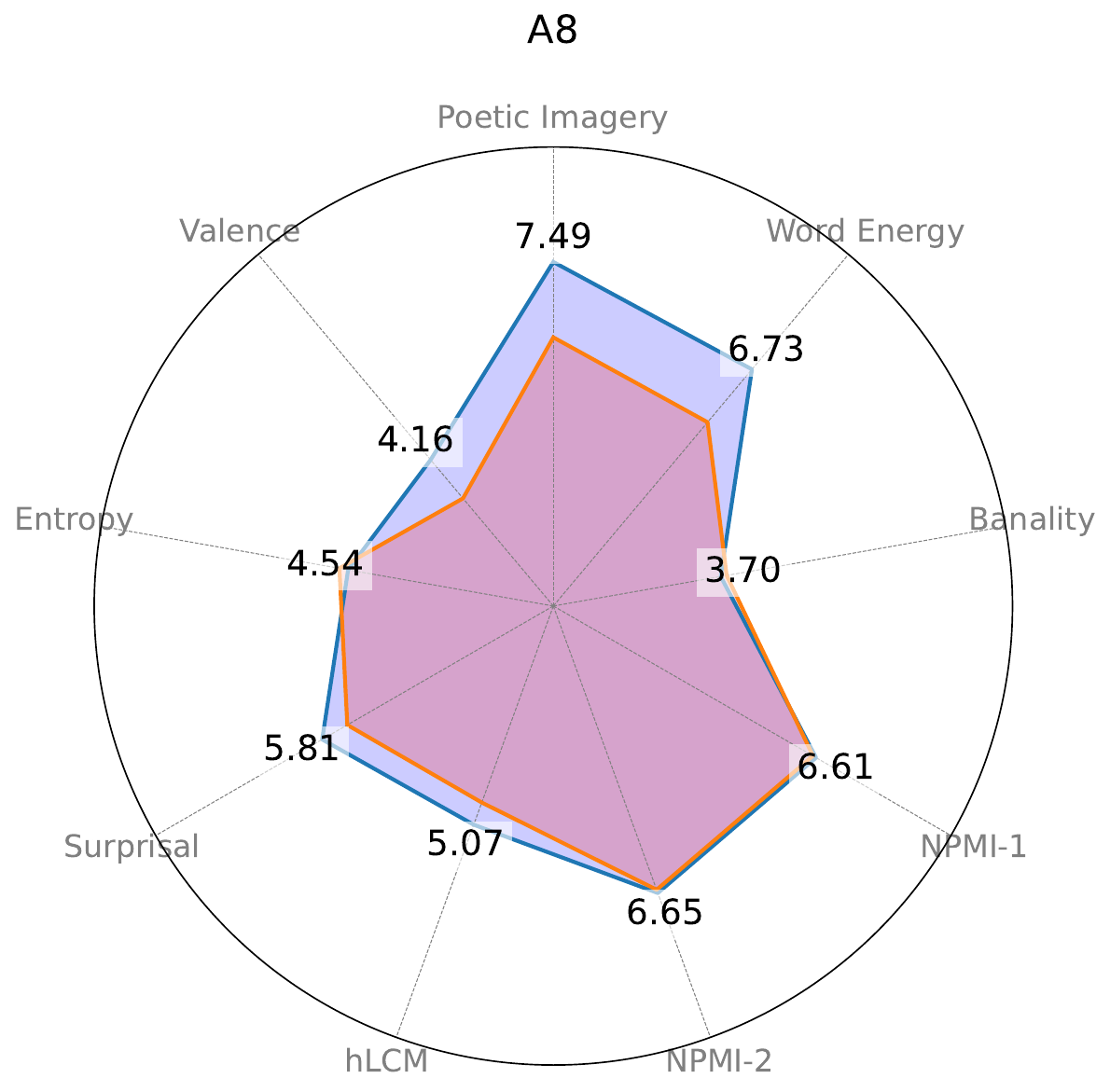} & \includegraphics[width=0.32\textwidth]{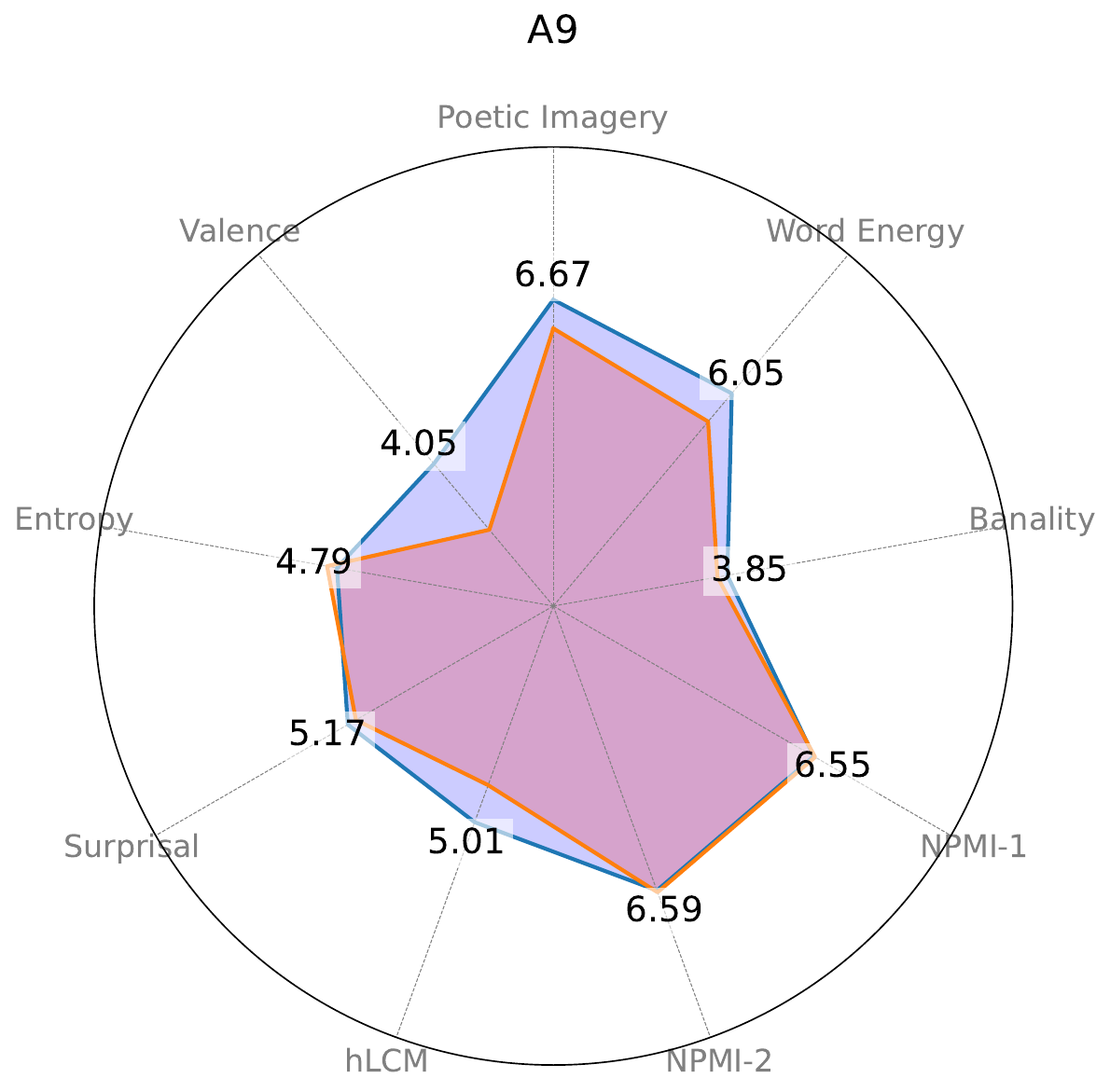} \\
& \includegraphics[width=0.3\textwidth]{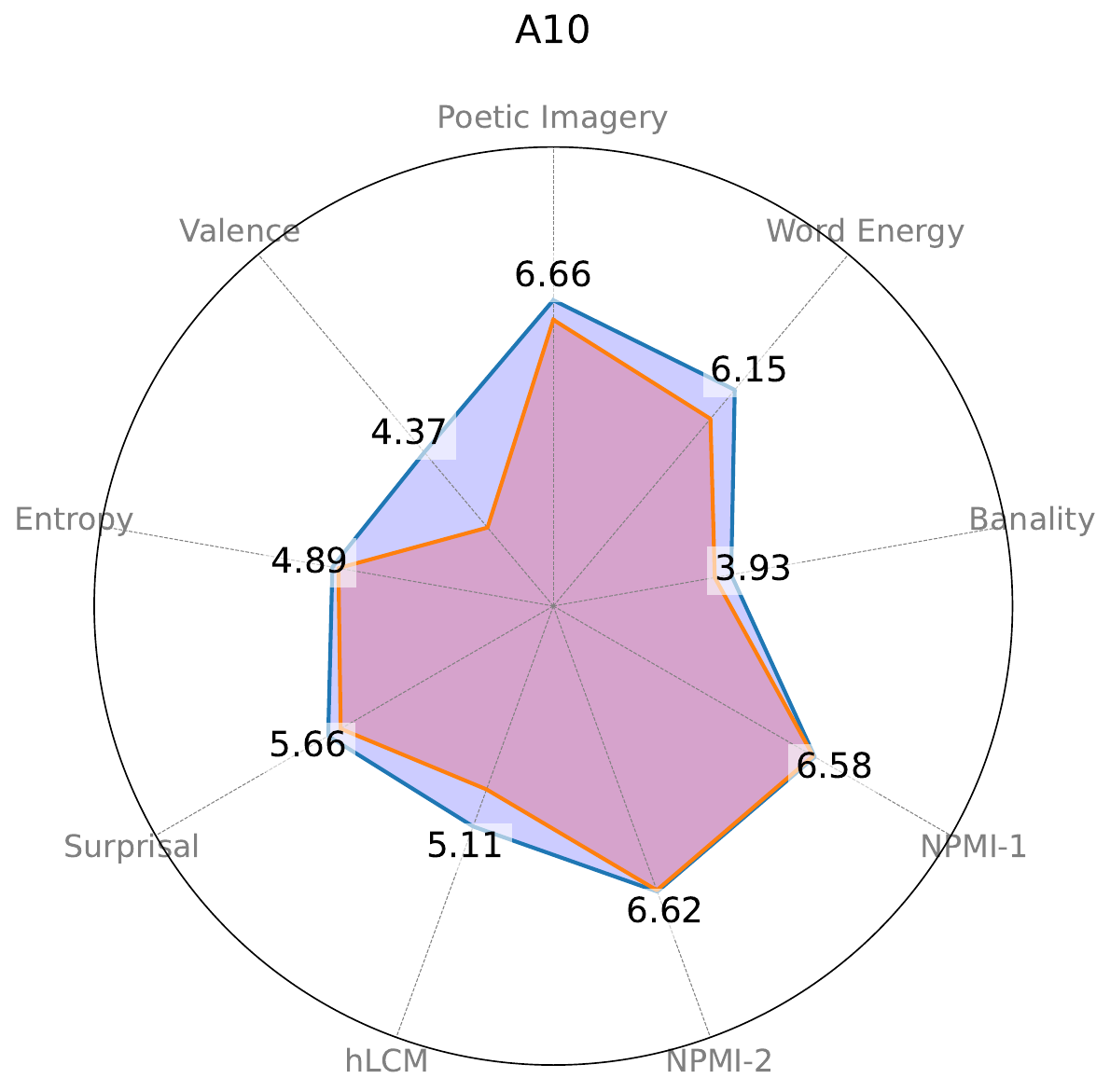} & \\
\end{tabular}}
\caption{Radar charts showing the 10 different {preference profiles of annotators} in the {\dataset} dataset. Blue webs denotes the web for positive lines, and red webs denotes the negative lines.}
\label{fig:profiles}
\end{figure*}

\section{Examples of Lyric Lines}
\label{app:web_lines}
\begin{figure*}[ht]
    \centering
    \includegraphics[width=0.85\textwidth]{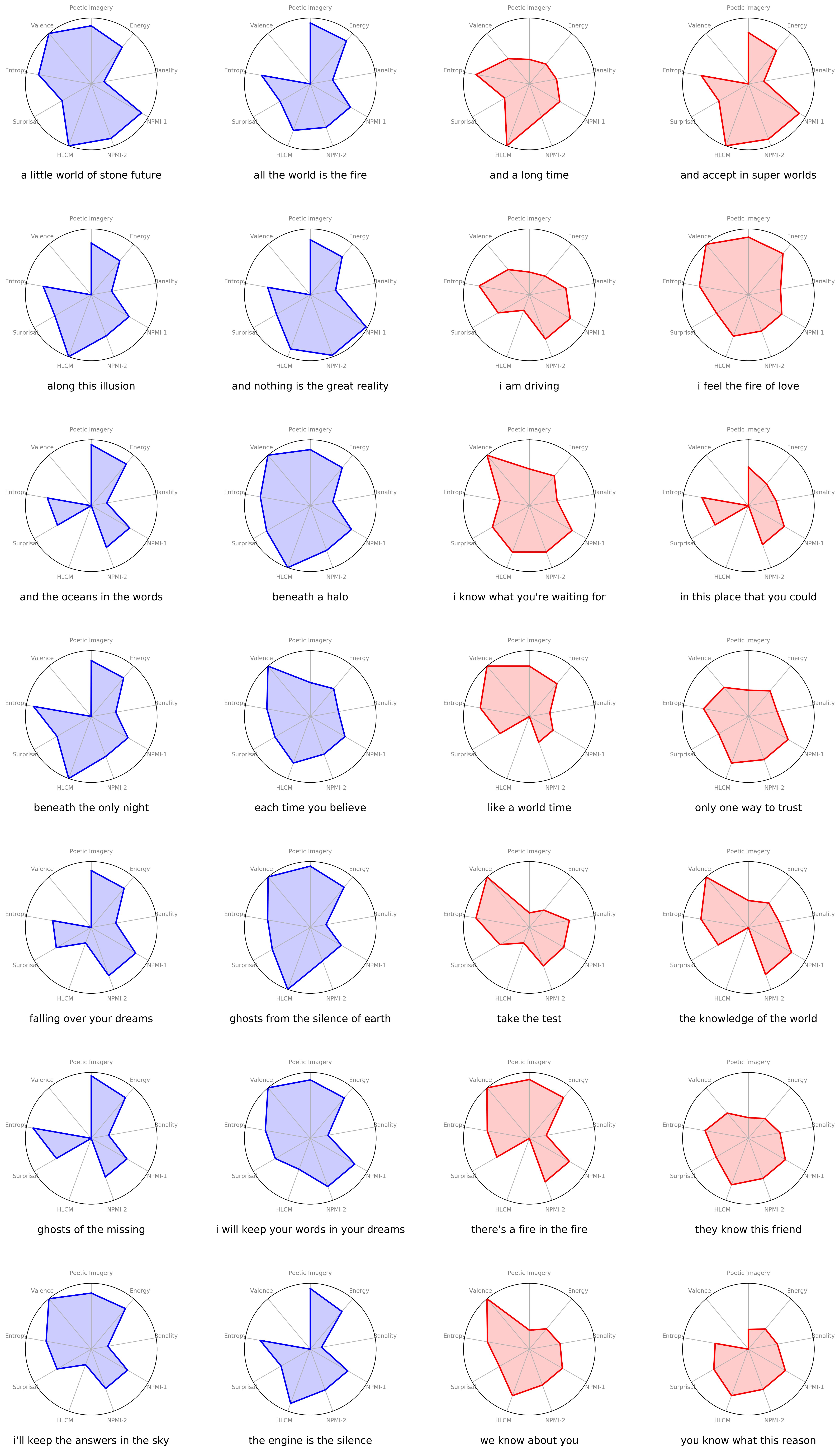}
    \caption{Examples of lyric lines rated as inspiring (blue) and not inspiring (red) by annotator $\mathcal{A}_1$.}
    \label{fig:web_diagram_grid_4x7_inverted_banality}
\end{figure*}

\section{Prompts Used for Various Tasks}
\label{app:prompts}
We only list some of our prompts due to space constraints but will release all the prompts upon acceptance.

% \begin{figure*}
%     \lstinputlisting[breaklines=true]{prompts/hLCM.txt}
%     \caption{Prompt used to obtain the number of occurrences of four linguistic categories (DAVs, IAVs, SVs, and ADJs) in a lyric line.}
%     \label{fig:hLCM}
% \end{figure*}

\begin{figure*}
\begin{lstlisting}[breaklines=true]
In this study, we aim to investigate the inspirational quality of song lyrics and poetic lines. Our objective is to understand how individual lyrics or poetic lines can influence a creative person. By ``inspiration," we refer to how these lines emotionally impact individuals and stimulate new creative ideas.

You will be shown 200 lines in total. Each line will be shown separately. Please take a moment to read each line and provide your rating based on its inspirational quality. We are interested in the potential of these lines to inspire creativity, rather than their grammatical correctness.

Please note that there are no right or wrong answers; we value your subjective impressions. Your feedback will help us understand how different lines of lyrics resonate with people.

The study is expected to take approximately 30 to 40 minutes. If you have to leave before you are done, you can resume the session using the same access code. Thank you for your participation!
\end{lstlisting}
    \caption{Guidelines presented to the eight human labelers.}
    \label{fig:user_study_txt}
\end{figure*}

% \begin{figure*}
%     \lstinputlisting[breaklines=true]{prompts/llm_classifier.txt}
%     \caption{Prompt used to classify a lyric line with LLaMA-3-70b.}
%     \label{fig:llm_clf}
% \end{figure*}

% \begin{figure*}
%     \lstinputlisting[breaklines=true]{prompts/valence_clf.txt}
%     \caption{Prompt used to classify a lyric line into one of 29 emotion categories with LLaMA-3-70b.}
%     \label{fig:valence_clf}
% \end{figure*}

\begin{figure*}
\begin{lstlisting}[breaklines=true]
PROMPT = """
You are a linguist who is analyzing the poetic imagery of a given textual expression.

### Instructions:
Given the expression:
"{sentence}"

### Task:
* Provide a rating between 1-5 on how much poetic imagery the expression has.
* You must enclose your rating between <rating></rating> tags.
* Do not generate anything else in your output.
* Follow the format below:
<rating>3</rating>

### Response:
"""
\end{lstlisting}
    \caption{Prompt used to measure poetic imagery of a lyric line.}
    \label{fig:pi_prompt}
\end{figure*}

% \begin{figure*}
%     \lstinputlisting[breaklines=true]{prompts/banality.txt}
%     \caption{Prompt used to measure banality of a lyric line.}
%     \label{fig:bnl_prompt}
% \end{figure*}

% \begin{figure*}
%     \lstinputlisting[breaklines=true]{prompts/word_energy.txt}
%     \caption{Prompt used to measure word energy of a lyric line.}
%     \label{fig:we_prompt}
% \end{figure*}

\end{document}